\documentclass[sigplan]{acmart}
\AtBeginDocument{%
  }

\setcopyright{acmlicensed}
\copyrightyear{2027}
\acmYear{2027}
\acmDOI{XXXXXXX.XXXXXXX}
\acmConference[ASPLOS 27]{}{June 03--05,
  2027}{Woodstock, NY}
\acmISBN{978-1-4503-XXXX-X/2018/06}

\usepackage{fancyhdr}
\usepackage[x11names]{xcolor}
\usepackage[ruled,vlined,linesnumbered]{algorithm2e}
\usepackage{multirow}
\usepackage{subfig}
\usepackage{graphicx}
\newcommand{\modelname}{\textit{VisMMOE }}
\newcommand{\modelnamenospace}{\textit{VisMMOE}}

\begin{document}

\title{VisMMOE: Exploiting Visual-Expert Affinity for Efficient Visual-Language MoE Offloading}

% %%
% %% The "author" command and its associated commands are used to define
% %% the authors and their affiliations.
% %% Of note is the shared affiliation of the first two authors, and the
% %% "authornote" and "authornotemark" commands
% %% used to denote shared contribution to the research.
% \affiliation{%

\author{Cheng Xu, Xiaofeng Hou$^{*}$, Jiacheng Liu, Chao Li$^{*}$}
\affiliation{%
  \institution{Shanghai Jiao Tong University}
  \city{Minhang Qu}
  \state{Shanghai}
  \country{China}
}

% 在作者后面紧跟这个命令来显示脚注
\thanks{$^{*}$Corresponding authors.}

% \affiliation{%

% \affiliation{%

% \affiliation{%


\begin{abstract}
% We present \modelnamenospace, a VL-MoE offloading system that improves expert locality and prefetch effectiveness through a compression-guided design. Our key observation, which we call \textit{visual-expert affinity}, is that removing redundant visual tokens not only reduces computation, but also makes expert activations more concentrated within layers and more stable across layers, yielding a smaller and more predictable expert working set. Based on this observation, \modelname combines three components: (1) affinity-aware token compression, which retains tokens that preserve both semantic fidelity and expert locality; (2) compression-guided expert prediction, which uses condensed visual signals to improve multi-layer lookahead; and (3) a multidimensional caching policy that mitigates offloading latency through pipelined prefetching.
% We implement \modelname on multiple frameworks and evaluate it on representative VL-MoE models and benchmarks. \modelname improves end-to-end inference performance by up to 2.68$\times$ and 1.61$\times$, respectively, over state-of-the-art baselines for today’s VL-MoE deployments.
Large-scale vision-language mixture-of-experts (VL-MoE) models provide strong multimodal capability, but efficient deployment on memory-constrained platforms remains difficult. Existing MoE offloading systems are largely designed for text-centric workloads and become much less effective for visual-heavy inputs, where large numbers of visual tokens induce broader and less predictable expert accesses.

We present \modelnamenospace, a VL-MoE offloading system built on a single systems insight: pruning redundant visual tokens can improve offloading not only by reducing computation, but also by reshaping expert demand. We refer to this effect as \textit{visual-expert affinity}: token pruning makes expert accesses more concentrated within layers and more stable across layers, producing a smaller and more predictable expert working set. Guided by this insight, \modelname combines affinity-aware token compression, lookahead expert prediction, and cache/pipeline orchestration to improve expert locality and prefetch effectiveness under tight memory budgets. We implement \modelname on multiple frameworks and evaluate it on representative VL-MoE models and benchmarks. \modelname improves end-to-end inference performance by up to 2.68$\times$ and 1.61$\times$, respectively, over strong baselines for today’s VL-MoE deployments while maintaining competitive accuracy.

\end{abstract}



\keywords{Multi-modal, Mixture-of-experts, Token compression, Expert cache, Prefetching}

\maketitle

\section{Introduction}

Vision-Language Large Language Models (VL-LLMs) are rapidly becoming a default interface for interactive AI, where each request may combine text with a large number of visual tokens from high-resolution images~\cite{wu2023multimodallargelanguagemodels,mmbenchiiswc}. To scale model capacity without proportionally increasing per-token computation, many VL-LLMs adopt Mixture-of-Experts (MoE)~\cite{moe,mxmoe,HybriMoE,expertflow,MoBiLE}, routing each token to only a small subset of experts. This sparsity makes MoE an attractive design point for large multimodal models such as MoE-LLaVA~\cite{moellava}, DeepSeek-VL2~\cite{deepseekvl2}, and Qwen-VL-MoE~\cite{Qwen3-VL}. Yet efficient single-GPU deployment remains challenging under tight memory budgets and latency constraints~\cite{10669603}, where expert movement from host to GPU often dominates runtime.

\begin{figure}[t]
\centering
    \includegraphics[width=\linewidth]{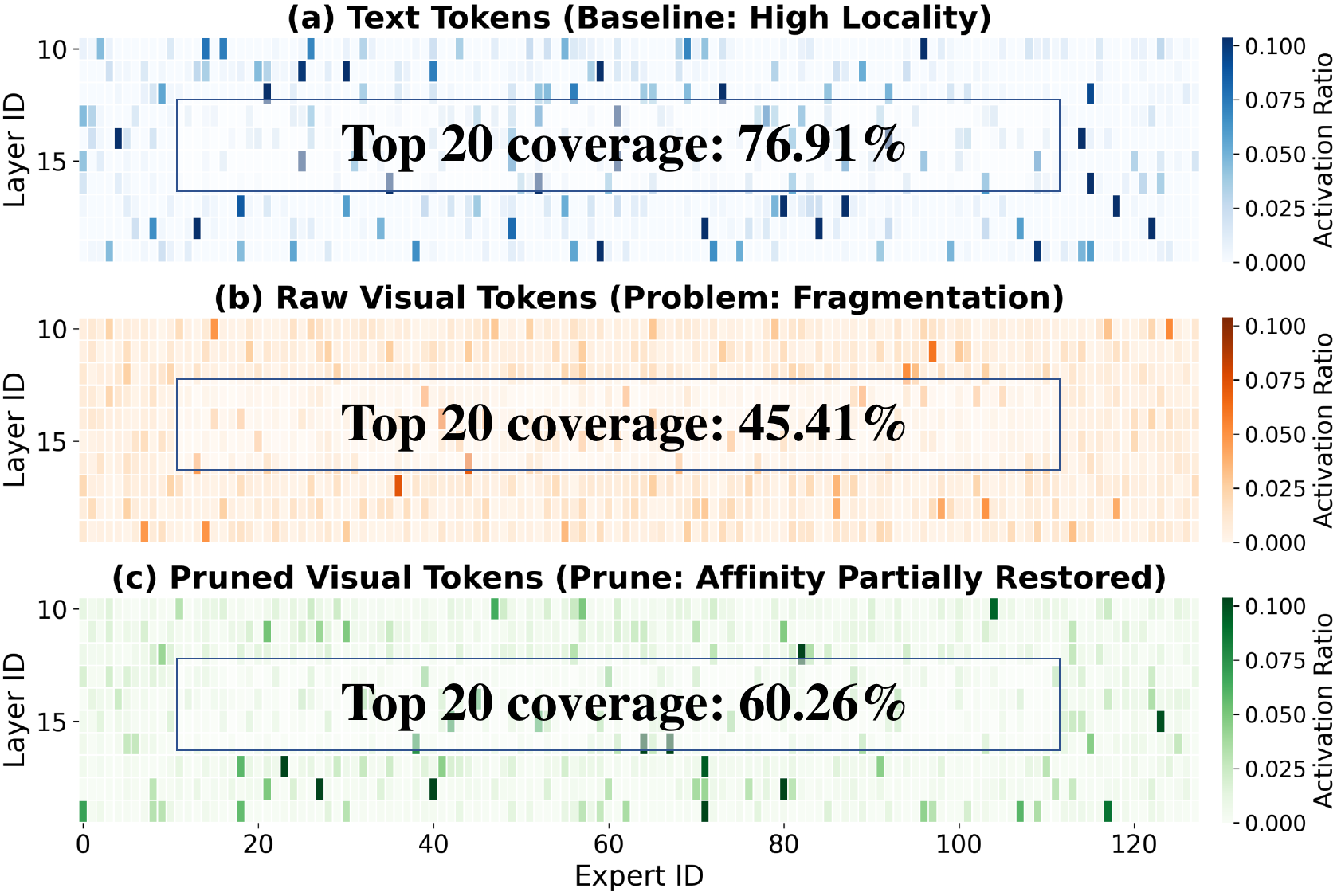}
    \caption{Expert activation distribution of Qwen3-VL-30B-A3B with 128 experts per layer. The x-axis is the expert ID, the y-axis is the layer ID, and darker color indicates a higher activation ratio. Text tokens exhibit strong locality, whereas raw visual tokens induce fragmented expert activations. After token pruning, the activation pattern becomes substantially more concentrated (+32\% relative).}
    \label{fig:intro}
\end{figure}

In practice, memory-constrained deployment~\cite{Latency-Memory-Trade-Off,MoBiLE,expertflow,SpecMD,cxl} often relies on \emph{expert offloading}, keeping only a small working set of experts in GPU memory while fetching the rest on demand. Prior MoE systems~\cite{hobbit,moe-offloading,moeinfinity} reduce swapping overhead through expert caching and predictive prefetching. These techniques~\cite{duoserve,song2025promoefastmoebasedllm} are effective when expert accesses remain concentrated and predictable, but they are substantially less effective for VL-MoEs. As shown in Figure~\ref{fig:intro}, text tokens activate a relatively compact subset of experts, whereas raw visual tokens induce much more scattered expert activations across layers. This fragmentation directly hurts offloading efficiency: it reduces cache reuse, increases non-reusable transfers, and often leaves expert movement on the critical path. Importantly, Figure~\ref{fig:intro} also shows that this behavior is not inherent to visual inputs. After pruning redundant visual tokens, the activation pattern becomes noticeably more concentrated, partially restoring locality and predictability.

This observation points to a systems opportunity rooted in the redundancy of visual inputs. Unlike text sequences, visual inputs contain many low-information patches that contribute little to downstream reasoning but can still perturb router decisions and activate a broad set of experts. While prior visual token compression methods primarily target compute reduction, our profiling reveals an additional systems effect: pruning redundant visual tokens can reduce the effective expert working set and improve routing regularity. We refer to this empirical trend as \textbf{Visual-Expert Affinity}. In VL-MoE deployment, token compression can therefore serve not only as a FLOPs-reduction tool, but also as a systems mechanism for improving cache reuse and prefetchability.
Exploiting this opportunity for efficient offloading, however, remains non-trivial.

\noindent\textbf{Challenge 1: token compression should be aligned with expert locality.}
Standard visual token compression is designed to preserve semantic salience and reduce FLOPs~\cite{visionzip,hiprune,xia2025fastmmoeacceleratingmultimodallarge}, but offloading efficiency is governed by a different objective: the size and compactness of the expert working set. As a result, tokens retained solely for semantic importance can still activate a dispersed set of experts, limiting parameter-access consolidation and keeping transfer overhead high. The runtime must therefore jointly balance reasoning fidelity and expert working-set compactness under GPU-memory and PCIe-bandwidth constraints.

\noindent\textbf{Challenge 2: offloading efficiency depends on accurate multi-layer prediction.}
Modern fine-grained MoEs substantially expand the routing search space. Compared with coarse-grained designs with 8--16 experts~\cite{hobbit,mixtral}, models such as Qwen-VL-MoE route over much larger expert pools (e.g., 128 experts)~\cite{Qwen3-VL}. In this regime, inaccurate prediction becomes increasingly expensive: prefetching too many irrelevant experts wastes bandwidth, while missing required experts leaves transfers on the critical path. Effective VL-MoE offloading therefore requires not only early lookahead, but also sufficiently precise prediction of the sparse expert working set needed in subsequent layers.

To address these challenges, we present \modelnamenospace, a VL-MoE offloading system that explicitly exploits Visual-Expert Affinity. \modelname combines \textbf{Affinity-Aware Token Compressor} with \textbf{Compression-Guided Lookahead Predictor} to (i) prune redundant tokens while compacting the expert working set, and (ii) predict future expert demand early enough to enlarge the prefetch window. \modelname further introduces \textbf{Expert Caching and Pipeline Orchestrator} that jointly coordinates computation, I/O, and memory capacity, while overlapping compression and prediction overhead with expert transfers whenever possible. We implement \modelname on commodity GPU platforms and evaluate it on representative VL-MoE models. Our results show that \modelname substantially reduces end-to-end inference latency over strong practical baselines while maintaining competitive accuracy.
In summary, this paper makes the following contributions:

\begin{itemize}
    \item We identify a VL-specific offloading mismatch: compared with text-centric MoE workloads, raw visual inputs induce broader and less predictable expert accesses, weakening both cache reuse and lookahead-based prefetching.

    \item We show that visual token compression can serve as a systems mechanism for offloading, not merely a compute optimization: pruning redundant visual tokens improves expert locality and routing regularity.

    \item Guided by this insight, we design a compression-guided VL-MoE offloading system that combines token retention, expert prediction, and runtime cache/pipeline management to reduce exposed transfer overhead under tight memory budgets.

    \item We implement \modelname on commodity GPU platforms and evaluate it across representative VL-MoE models, runtimes, and hardware targets, demonstrating substantial end-to-end latency reductions over strong practical baselines.
\end{itemize}

The rest of the paper is organized as follows. Section~\ref{sec:background} introduces the background and motivation. Section~\ref{sec:system} presents the system design. Section~\ref{sec:evaluation} evaluates the system. Section~\ref{sec:relatedworks} discusses related work, and Section~\ref{sec:conclusion} concludes.

\begin{figure}[t]
\centering
    \includegraphics[width=\linewidth]{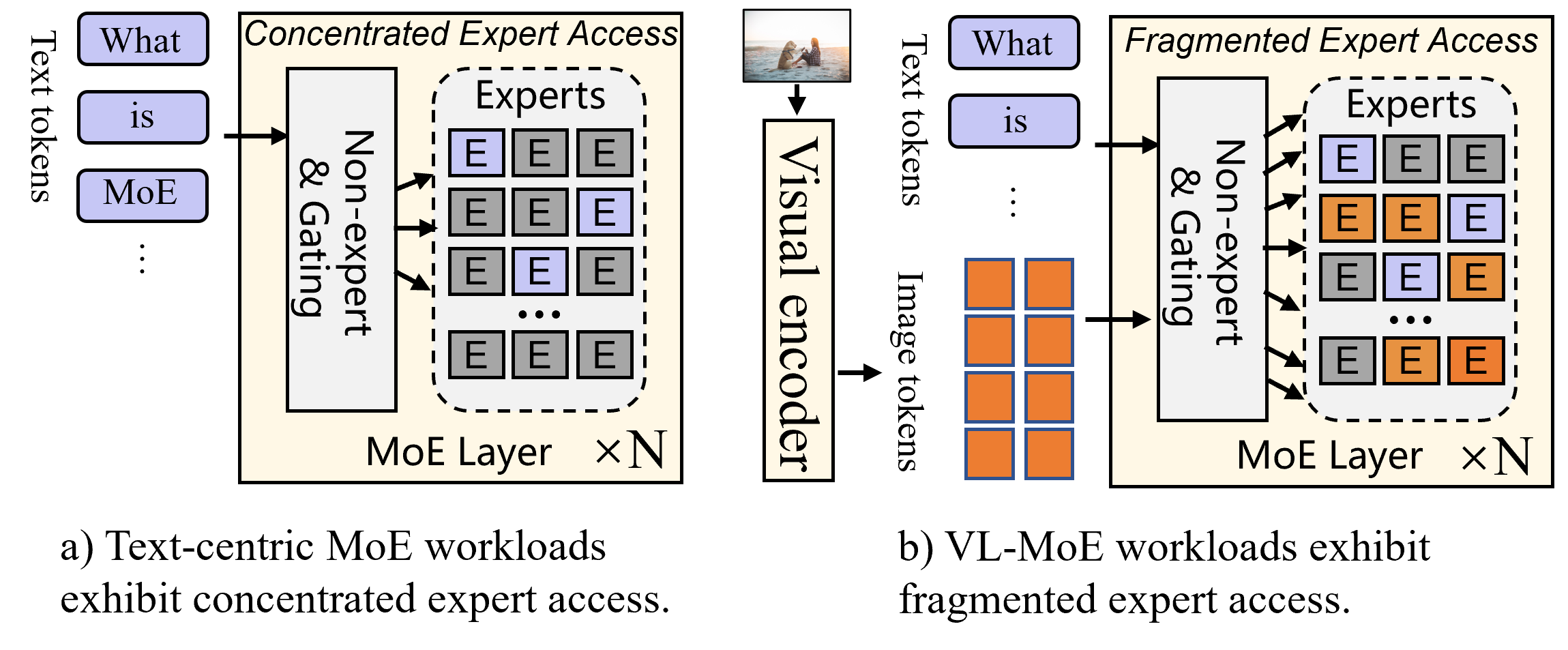}
    \caption{Why prior MoE offloading assumptions break for VL-MoEs: text-centric workloads induce concentrated expert access, whereas visual-heavy VL-MoE inputs broaden the effective expert working set.}
    \label{fig:background}
\end{figure}

\section{Background and Motivation}
\label{sec:background}

\subsection{VL-MoE Models and GPU Offloading}
VL-MoE models combine a visual encoder with a Transformer language backbone whose feed-forward blocks are replaced by sparse MoE layers~\cite{Qwen3-VL,deepseekvl2,prism}. Similar to text-only MoE models, each token activates only a small subset of experts in each layer, reducing per-token computation while preserving large model capacity. However, the full expert pool remains too large to fit on a single GPU and must therefore be managed through expert offloading.

Existing MoE offloading systems~\cite{hobbit,moeinfinity,song2025promoefastmoebasedllm,similaritymeasure,spac} rely on a key assumption: expert accesses are sufficiently predictable that caching and prefetching can effectively hide data movement. This assumption is often reasonable in text-centric MoE workloads, where tokens repeatedly activate a compact subset of experts. As illustrated in Figure~\ref{fig:background}(a), such concentrated access leads to a small and reusable expert working set that is favorable for GPU caching and predictive prefetching.

VL-MoEs, however, operate in a different regime~\cite{modes,xia2025fastmmoeacceleratingmultimodallarge}. In addition to text, each request may contain a large number of visual patch tokens produced by the visual encoder. These visual tokens induce broader and less regular routing behavior than text-centric workloads, expanding the expert working set and weakening routing predictability. As shown in Figure~\ref{fig:background}(b), VL-MoE requests therefore violate the locality assumptions that prior offloading strategies depend on.

% \begin{figure}[t]
% \centering
%     \includegraphics[width=\linewidth]{figures/io-motivation.pdf}
%     \vspace{-0.7cm}
%     \caption{Latency breakdown of a standard VL-MOE layer. Full loads means loading all experts (128 for Qwen and 72 for DeepSeek-VL2), while predictive load predictively load 20 experts. Even predictive IO far overweigh other operations.} 
%     \label{fig:io-motivation}
%     \vspace{-0.3cm}
% \end{figure}

\begin{table}[t]
\centering
\caption{Analysis of Visual vs. Textual token distribution.}
\label{tab:token_stats}
\resizebox{\linewidth}{!}{
\begin{tabular}{lcccc}
\toprule
\textbf{Dataset} & \textbf{Visual} & \textbf{User Text} & \textbf{Total Text}$^{\dagger}$ & \textbf{Ratio (V/T)} \\
\midrule
MMBench   & 197  & 11 & 53 & 3.72$\times$ \\
% OCRBench  & 261  & 7  & 49 & 5.33$\times$ \\
MME       & 1107 & 18 & 60 & \textbf{18.45$\times$} \\
POPE      & 358  & 8  & 50 & 7.16$\times$ \\
\bottomrule
\multicolumn{5}{l}{\footnotesize $^{\dagger}$Includes a fixed 42-token system prompt overhead in this model.}
\end{tabular}
}
\end{table}

\subsection{Characterizing Raw VL-MoE Offloading}

Expert movement is often the dominant bottleneck in memory-constrained MoE offloading. In our VL-MoE setting, even with predictive loading of 20 experts, CPU--GPU transfer still accounts for 62.1\% and 86.9\% of layer latency for Qwen3-VL and DeepSeek-VL2, respectively, indicating that expert movement frequently remains on the critical path (detailed in \ref{appendix:io}). However, VL-MoEs are challenging not merely because offloading is I/O-bound in general, but because visual-heavy inputs break the locality and predictability assumptions that prior MoE offloading systems rely on~\cite{hobbit,moeinfinity,song2025promoefastmoebasedllm}. We characterize this mismatch from two complementary perspectives.

\noindent\textbf{Problem 1: Visual-heavy token composition.}
VL-MoE requests are dominated by image tokens rather than text tokens. As shown in Table~\ref{tab:token_stats}, visual tokens outnumber text tokens by 3.7$\times$ to 18.4$\times$ across representative multimodal benchmarks. Thus, VL-MoE inference is not simply text-centric MoE with a small visual prefix; it operates in a visual-heavy regime, where expert offloading pressure is substantially amplified by the large number of visual tokens.

\noindent\textbf{Problem 2: Fragmented expert demand.}
Token volume alone does not fully explain the difficulty. The more fundamental issue is that raw visual inputs broaden the effective expert working set and reduce routing regularity across layers. As illustrated in Figure~\ref{fig:intro} and ~\ref{fig:background}, expert accesses under raw visual inputs span a much larger subset of experts than text-centric inputs. This lower locality directly reduces cache residency effectiveness and makes future expert demand harder to predict for lookahead-based offloading.

The next subsection shows that pruning redundant visual tokens helps precisely along these two dimensions: it spatially compacts expert demand within each layer and temporally stabilizes routing behavior across layers, directly addressing both problems above.

% \begin{figure}[t]
%     \centering
%     \subfloat[Expert-access concentration.\label{fig:motivation-affinity-a}]{
%         \includegraphics[width=0.47\linewidth]{figures/fig4a.png}
%     }
%     \hfill
%     \subfloat[Latency breakdown.\label{fig:motivation-affinity-b}]{
%         \includegraphics[width=0.47\linewidth]{figures/fig4b.png}
%     }
%     \vspace{-0.4cm}
%     \caption{Spatial impact of token pruning. Pruning redundant visual tokens increases expert-access concentration and, even without prefetching, lowers prefill offloading cost by shrinking the set of experts that must be loaded on demand.}
%     \label{fig:motivation-affinity}
%     \vspace{-0.4cm}
% \end{figure}

% in preamble

\begin{figure}[t]
    \centering
    \subfloat[Expert-access concentration.\label{fig:motivation-affinity-a}]{
        \includegraphics[width=0.47\linewidth]{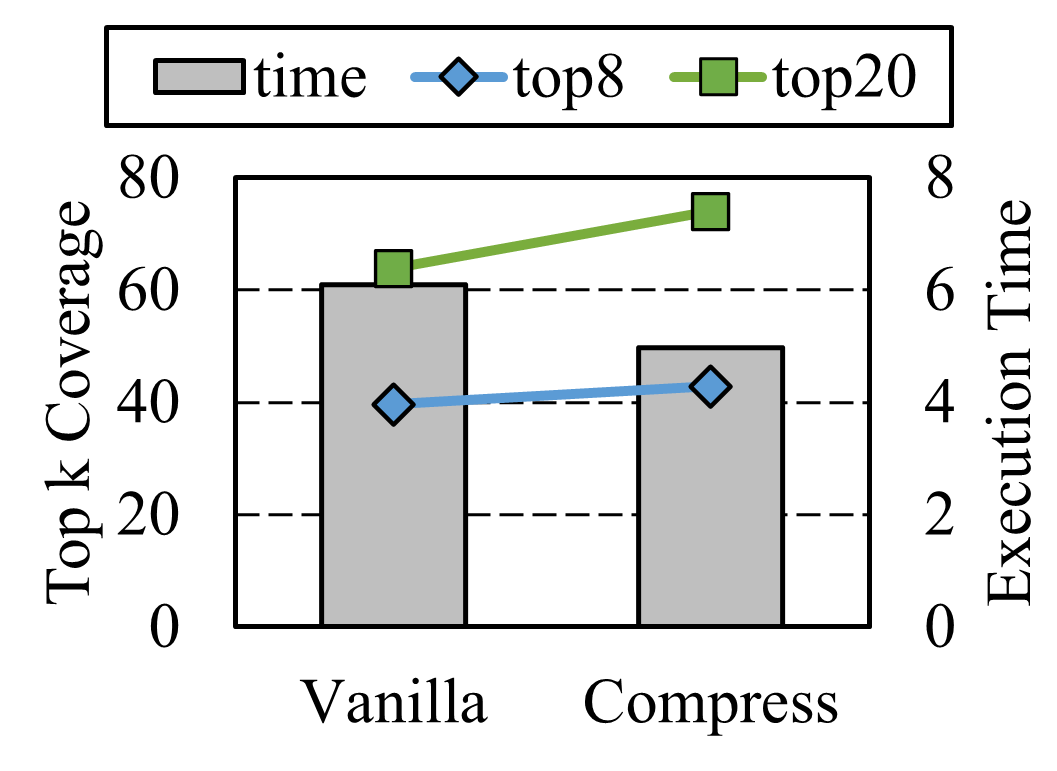}
    }
    \hfill
    \subfloat[Working-set effects of pruning.\label{fig:motivation-affinity-b}]{
        \includegraphics[width=0.47\linewidth]{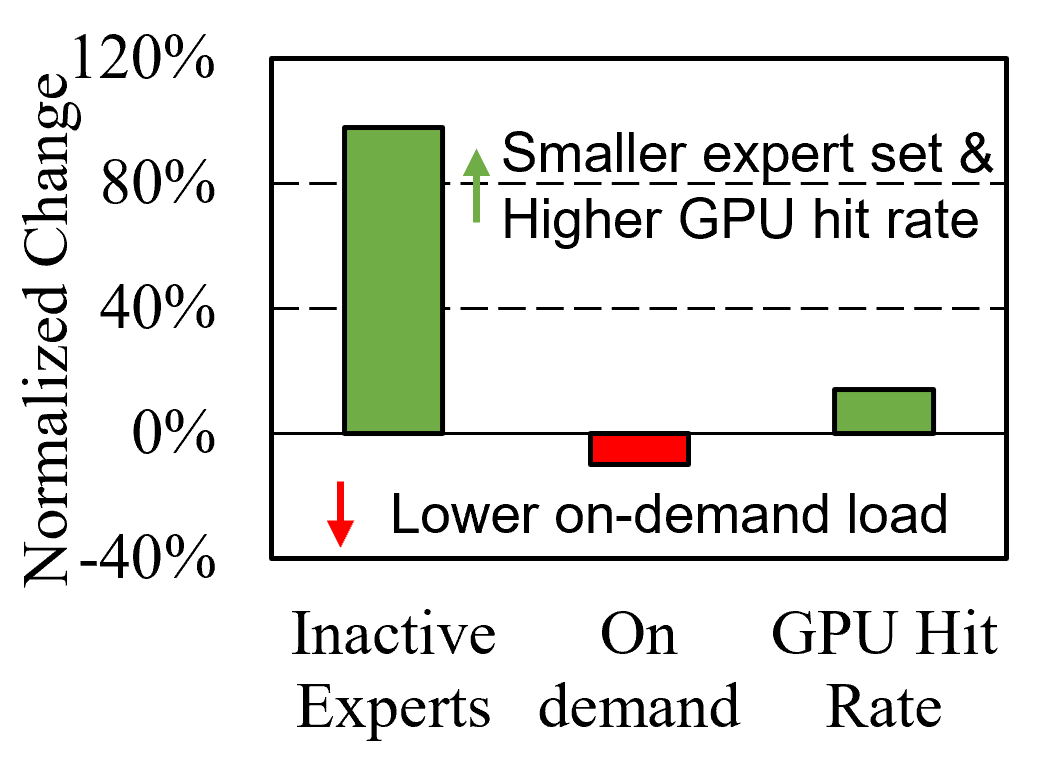}
    }
    \caption{Spatial impact of token pruning. (a) Pruning redundant visual tokens increases expert-access concentration, improves top-$k$ expert coverage and thus reduces prefill time. (b) This concentration translates into a smaller and more reusable expert working set: the number of inactive experts nearly doubles, on-demand expert loads decrease, and GPU expert hit rate improves during prefill.}
    \label{fig:motivation-affinity}
\end{figure}

\begin{figure}[t]
\centering
    \includegraphics[width=\linewidth]{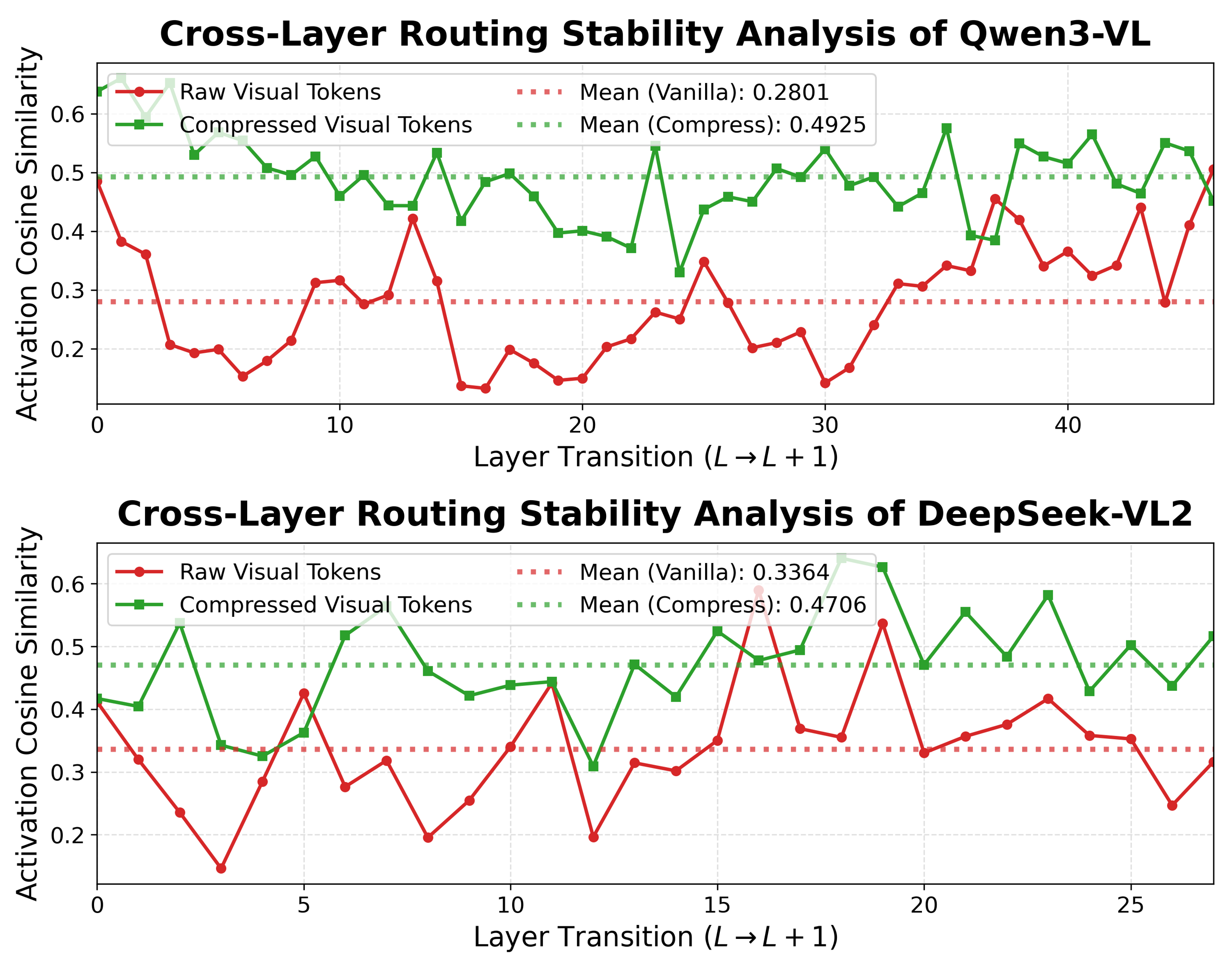}
    \caption{Temporal impact of token pruning. Pruning redundant visual tokens increases inter-layer routing similarity, making future expert demand more predictable for lookahead-based prefetching.}
    \label{fig:stability}
\end{figure}

\begin{figure*}[t]
\centering
    \includegraphics[width=\linewidth]{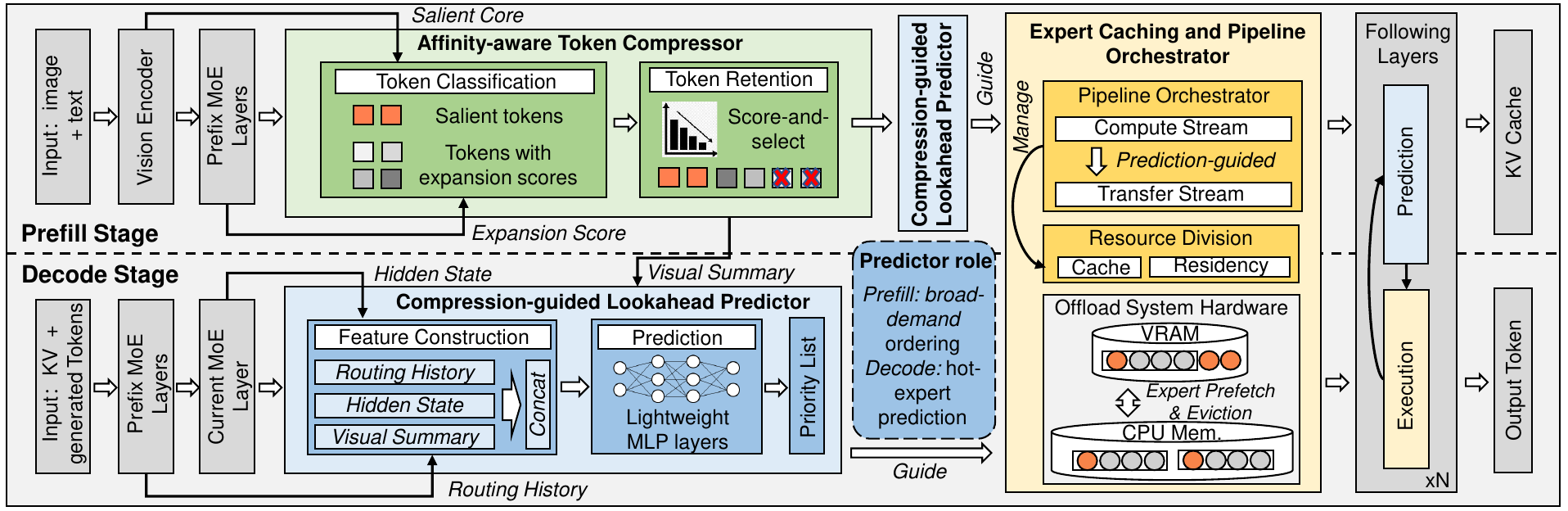}
    \caption{Overview of the \modelname architecture. In prefill, raw multimodal inputs are compressed by the \textit{Affinity-aware Token Compressor}. The \textit{Compression-guided Lookahead Predictor} produces phase-dependent priority signals: broad-demand ordering in prefill and hot-expert prediction in decode. Guided by these signals, the \textit{Expert Caching and Pipeline Orchestrator} manages cache residency and overlaps expert transfer with execution.}
    \label{fig:overview}
    \vspace{-0.2cm}
\end{figure*}

\subsection{Opportunity: Visual-Expert Affinity}

Recent visual token compression methods primarily target compute reduction and KV-cache savings~\cite{visionzip,hiprune,LLaVA-Scissor}. Our profiling reveals an additional systems opportunity for VL-MoE offloading: pruning redundant visual tokens also reshapes expert demand in ways that are directly beneficial for data movement. We refer to this empirical trend as \textbf{Visual-Expert Affinity}. Rather than viewing token compression solely as a FLOPs optimization, we observe that it can also improve the locality and predictability properties that efficient expert offloading depends on.

\noindent\textbf{Spatial concentration.}
Token pruning improves the spatial compactness of expert demand within each layer. Many low-information visual patches contribute little to downstream reasoning, yet still inject noisy routing signals and activate a broad set of experts. As shown in Figure~\ref{fig:motivation-affinity}(a), pruning such redundant tokens increases top-$k$ expert coverage and concentrates routing onto a smaller expert subset. This concentration shrinks the effective on-demand expert set: Figure~\ref{fig:motivation-affinity}(b) shows that more experts remain inactive, on-demand expert loads decrease, and GPU expert hit rate improves during prefill. Thus, even without prediction or prefetching, spatial concentration alone already lowers prefill offloading overhead.

\noindent\textbf{Temporal stability.}
Token pruning also improves routing regularity across layers. Owing to the residual structure of Transformer backbones, hidden states---and thus MoE gating inputs---are naturally similar across nearby layers. Raw visual tokens, however, inject noisy and fragmented routing signals that weaken this cross-layer consistency and make future expert demand harder to anticipate. As shown in Figure~\ref{fig:stability}, raw visual tokens exhibit lower inter-layer routing similarity than text-centric inputs. After pruning, the retained tokens induce more stable gating trajectories across consecutive layers. This increased temporal stability makes the expert working set of subsequent layers more predictable, thereby improving the effectiveness of lookahead-based prefetching.

Together, these observations show that visual token compression can serve not only as a compute reduction, but also as a systems optimizer in offloading. It motivates a design that aligns token retention with expert locality and future expert demand, which we present in the next section.

\section{\modelname System}
\label{sec:system}

\subsection{\modelname Overview} 

The challenges and opportunities discussed in Section~\ref{sec:background} highlight the need for a VL-MoE offloading design that goes beyond simplistic per-layer prefetching. To this end, we present \modelnamenospace, an offloading system that explicitly leverages Visual-Expert Affinity. By compacting expert demand, enabling deeper lookahead prediction, and coordinating data movement, \modelname reduces exposed PCIe transfer overhead and improves end-to-end inference efficiency under tight memory budgets.
Figure~\ref{fig:overview} illustrates the architecture of \modelnamenospace, which consists of three components for efficient VL-MoE offloading:

\begin{itemize}
    \item \textbf{Affinity-aware Token Compressor.} This module retains visual tokens according to semantic importance and expert affinity. It reduces the effective expert working set and makes expert activations more concentrated, improving both transfer efficiency and cross-layer predictability.
    
    \item \textbf{Compression-Guided Lookahead Predictor.} Using the condensed visual signals from the compression module, this module estimates the experts required in future layers, enabling lookahead for prefetching.
    
    \item \textbf{Expert Caching and Pipeline Orchestrator.} Based on the predicted expert working sets, this module coordinates GPU memory, including a small resident prefix for routing observability, PCIe transfers, and model execution so that expert movement can be largely overlapped with computation whenever possible.
\end{itemize}

The three components of \modelname are tightly coupled.
Affinity-aware compression reshapes expert demand, enabling more accurate prediction and more effective cache/pipeline orchestration.
Without this demand-shaping effect, both prediction and cache reuse degrade under visual-heavy inputs.

\subsection{Affinity-aware Token Compressor}

To operationalize \textit{Visual-Expert Affinity}, we design \textbf{Affinity-aware Token Compressor}, a visual-token selection policy tailored for VL-MoE offloading as shown in Algorithm~\ref{alg:affinity_pruning}. Unlike prior compression methods that primarily optimize FLOPs or KV-cache usage, our objective is explicitly system-facing: under a fixed keep budget, retain tokens that preserve useful visual evidence while limiting unnecessary expansion of the expert working set. Intuitively, once the indispensable visual evidence has been preserved, the remaining token budget should not simply be spent on the next-most-salient tokens in isolation; it should be spent on tokens that add useful context with minimal additional expert demand.
Formally, let $V=\{v_i\}_{i=1}^{N}$ denote the visual tokens, and let $K_{\text{keep}}=\lfloor \beta N \rfloor$ be the total keep budget under keep ratio $\beta$. Our design proceeds in two steps. We first identify a salient core that serves as an accuracy anchor, and then allocate the remaining budget using a calibrated working-set-aware score.

\begin{algorithm}[t]
\small
\SetAlgoLined
\DontPrintSemicolon
\SetKwInOut{Input}{Input}
\SetKwInOut{Output}{Output}
\SetKwFunction{TopK}{TopK}
\SetKwFunction{Mean}{Mean}
\SetKwFunction{Active}{ActiveExperts}
\SetKwFunction{Normalize}{Normalize}
\definecolor{stepColor}{RGB}{0, 51, 153}
\definecolor{commColor}{RGB}{85, 107, 47}

\newcommand{\AlgoStep}[1]{\vspace{2pt}\tcc{\textcolor{stepColor}{\textbf{#1}}}}
\newcommand\mycommfont[1]{\footnotesize\ttfamily\textcolor{stepColor}{#1}}
\SetCommentSty{mycommfont}
\caption{Affinity-Aware Token Compression}
\label{alg:affinity_pruning}

\Input{Attention maps $\mathbf{A}$, prefix routing outputs $\mathbf{R}$, salient ratio $\alpha$, keep ratio $\beta$, trade-off weight $\lambda$}
\Output{Retained token indices $\mathcal{I}_{\text{keep}}$}

$N \leftarrow |\mathbf{A}|$\;
$\mathbf{s} \leftarrow \Mean_h(\mathbf{A}^h)$ \tcp*[r]{Head-averaged token saliency}
$\tilde{\mathbf{s}} \leftarrow \Normalize(\mathbf{s})$ \tcp*[r]{Normalize to $[0,1]$}

$K_{\text{core}} \leftarrow \lfloor \alpha N \rfloor$,\quad
$K_{\text{keep}} \leftarrow \lfloor \beta N \rfloor$,\quad
$K_{\text{rem}} \leftarrow K_{\text{keep}} - K_{\text{core}}$\;

$\mathcal{I}_{\text{core}} \leftarrow \TopK(\tilde{\mathbf{s}}, K_{\text{core}})$ \tcp*[r]{Salient core}
$\mathcal{E}_{\text{target}} \leftarrow \bigcup_{j \in \mathcal{I}_{\text{core}}} \Active(\mathbf{R}_j)$ \tcp*[r]{Core-induced expert set}

\ForEach{$i \notin \mathcal{I}_{\text{core}}$}{
    $\mathcal{E}_i \leftarrow \Active(\mathbf{R}_i)$\;
    $\Delta_i \leftarrow |\mathcal{E}_i \setminus \mathcal{E}_{\text{target}}| / |\mathcal{E}_i|$ \tcp*[r]{Marginal expansion}
    $p_i \leftarrow \tilde{s}_i - \lambda \Delta_i$ \tcp*[r]{Working-set-aware score}
}

$\mathcal{I}_{\text{extra}} \leftarrow \TopK(\mathbf{p}, K_{\text{rem}})$,
$\mathcal{I}_{\text{keep}} \leftarrow \mathcal{I}_{\text{core}} \cup \mathcal{I}_{\text{extra}}$\;

\Return{$\mathcal{I}_{\text{keep}}$}
\end{algorithm}

\subsubsection{Salient Core as an Accuracy Anchor}

We first identify a small set of visually indispensable tokens that anchors semantic fidelity. Importantly, this \emph{salient core} is not the final compressed set; rather, it provides an accuracy-preserving foundation on top of which the remaining keep budget is allocated. Keeping only the most salient tokens is often too aggressive for downstream VL reasoning, while retaining additional context remains beneficial as long as it does not excessively expand expert demand.

Following standard visual token compression practice~\cite{visionzip,hiprune}, we estimate token saliency from the attention maps of the vision encoder. Let $A$ denote the attention tensor and let $s_i$ be the head-averaged saliency score of token $v_i$. We then normalize saliency scores to $\tilde{s}_i \in [0,1]$ and retain the top $K_{\text{core}}=\lfloor \alpha N \rfloor$ tokens as the salient core:
\begin{equation}
I_{\text{core}} = \operatorname{TopK}(\tilde{s}, K_{\text{core}}),
\end{equation}
where $\alpha$ is a small salient ratio. These tokens preserve the most important visual evidence and define the minimum semantic context that should not be discarded.

The salient core also induces an \emph{unavoidable} expert set. Specifically, we use router outputs from the pinned prefix of early MoE layers and, for each token, collect the experts it activates across those layers; their union defines the token's active expert set for affinity scoring. The expert set induced by the core tokens therefore captures the prefix-scoped expert working set already required to preserve the visual semantics. The remaining compression problem is thus not simply which tokens are salient in isolation, but which additional tokens provide useful context with minimal expansion cost. This prefix-scoped expert representation is used only for compression-time affinity scoring and is distinct from the predictor output space in Section~\ref{subsec:prediction}.

\subsubsection{Calibrated Marginal Expansion Score}

Given the salient core, we allocate the remaining token budget using a \textbf{calibrated marginal expansion score}. The goal is to retain additional tokens that preserve useful visual context while introducing as little new expert demand as possible beyond the core-induced expert set.

For each non-core token $v_i$, let $E_i=\operatorname{ActiveExperts}(R_i)$ denote the experts activated by that token. We define its marginal expansion score relative to the salient core as
\begin{equation}
\Delta_i = \frac{|E_i \setminus E_{\text{target}}|}{|E_i|}.
\end{equation}
Equivalently, if $m_i \in \{0,1\}^{E}$ and $m_{\text{target}} \in \{0,1\}^{E}$ denote the corresponding multi-hot expert masks, then
\begin{equation}
\Delta_i = \frac{\|m_i \odot (1-m_{\text{target}})\|_1}{\|m_i\|_1}.
\end{equation}
This score measures how much new expert demand a token would introduce beyond the current core-induced working set. Tokens that largely reuse the core-induced experts incur low marginal expansion, while tokens that introduce many new experts incur high marginal expansion. 
% In implementation, these set differences are computed with compact multi-hot masks and bitwise operations rather than explicit variable-length set intersections.

Based on this score, we rank non-core tokens by
\begin{equation}
p_i = \tilde{s}_i - \lambda \Delta_i
\end{equation}
where $\lambda$ balances semantic value against working-set growth. We use a fixed global value of $\lambda = 2$ throughout the evaluation since \ref{appendix:lambda_sensitivity} shows that this setting captures most of the latency benefit while preserving competitive accuracy.
Since both $\tilde{s}_i$ and $\Delta_i$ are normalized to $[0,1]$, this value assigns twice the weight to working-set expansion as to semantic saliency, reflecting the I/O-dominated nature of expert offloading. This single value is shared across all models and benchmarks.
 The final retained set is
\begin{equation}
I_{\text{keep}} = I_{\text{core}} \cup \mathrm{TopK}(\{p_i\}_{i \notin I_{\text{core}}},\, K_{\text{keep}} - K_{\text{core}}).
\end{equation}
This score-guided allocation explains why we do not stop at the top-$\alpha N$ salient tokens. The salient core anchors accuracy, while the remaining budget preserves additional context in a working-set-aware manner, yielding a compressed set that retains more useful visual evidence than core-only pruning and induces a smaller, more reusable expert working set than saliency-only retention. Since affinity-aware selection requires routing signals unavailable from raw visual inputs, we derive them from a small pinned prefix of early MoE layers on GPU, whose router outputs expose the expert structure needed for scoring; its broader runtime role is discussed in Section~\ref{subsec:optimization}.

\subsection{Compression-guided Lookahead Predictor}
\label{subsec:prediction}
\begin{figure}[t]
\centering
    \includegraphics[width=\linewidth]{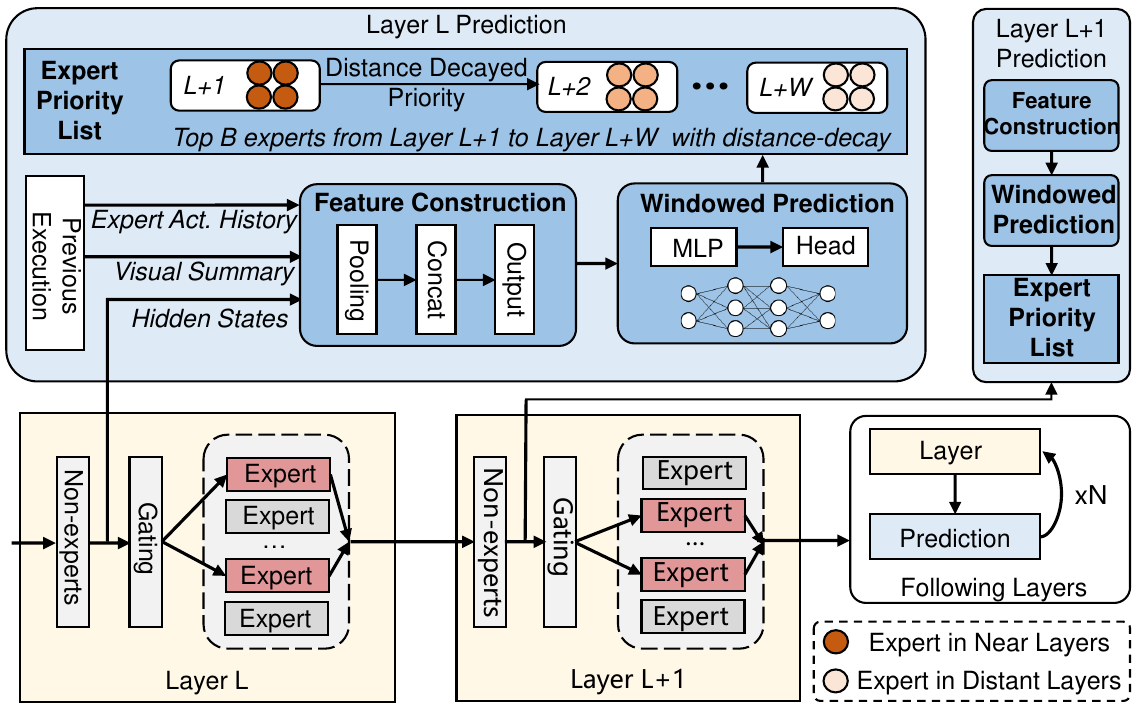}
    \caption{Architecture of the compression-guided lookahead predictor. The predictor pools compressed hidden-state and visual summaries from the retained token set, combines them with routing history, and predicts a priority score over experts within a bounded lookahead window.}
    \label{fig:predictor}
    \vspace{-0.2cm}
\end{figure}

We introduce a \textbf{compression-guided lookahead predictor} that runs alongside the main execution loop (Figure~\ref{fig:predictor}) and provides an early priority signal for near-future expert demand. In \textbf{prefill}, where demand is broader, this signal serves mainly as a bounded priority ordering under limited VRAM and bandwidth: higher-scored experts are fetched earlier, improving overlap and cache residency, while experts absent from the realized routes need not be fetched. In \textbf{decode}, where demand is much sparser and more reusable, the same signal more directly identifies near-future hot experts and avoids unnecessary prefetches.

\subsubsection{Compression-aware Feature Construction}

At MoE layer $l$, the predictor forms a compact feature vector $\mathbf{x}_l$ from three signals:
(i) \textbf{routing history} $\mathbf{h}^{(r)}_l \in \mathbb{R}^{D_r}$, a compressed histogram of recent expert activations;
(ii) \textbf{compressed hidden-state summary} $\mathbf{h}^{(h)}_l \in \mathbb{R}^{D_h}$, obtained by pooling only over the retained visual tokens from Section~3.2; and
(iii) \textbf{compressed visual summary} $\mathbf{h}^{(v)} \in \mathbb{R}^{D_v}$, a static visual descriptor computed from the same retained token set.
Let $\mathcal{I}_{\text{keep}}$ denote the retained visual-token indices after affinity-aware compression. We define:

{\small
\begin{equation}
\mathbf{h}^{(h)}_l = \mathrm{MP}(\{\mathbf{h}_{l,j}: j \in \mathcal{I}_{\text{keep}}\}), \quad
\mathbf{h}^{(v)} = \mathrm{MP}(\{\mathbf{v}_j: j \in \mathcal{I}_{\text{keep}}\}).
\end{equation}
}
where $\mathbf{h}_{l,j}$ is the hidden state of retained token $j$ at layer $l$, and $\mathbf{v}_j$ is its corresponding visual embedding. By mean pooling (MP) only over the affinity-aware compressed tokens, the predictor both reduces its feature-processing cost and suppresses noisy visual patches. The final predictor input is
{\small
\begin{equation}
\mathbf{x}_l = [\mathbf{h}^{(r)}_l; \mathbf{h}^{(h)}_l; \mathbf{h}^{(v)}].
\end{equation}
}
During autoregressive decoding, repeatedly reading the retained visual hidden states from the KV-cache would incur prohibitive memory bandwidth. Therefore, we reuse the static visual summary \(\mathbf{h}^{(v)}\) throughout decode. 

\subsubsection{Windowed Prediction with Decayed Supervision}

Rather than predicting expert usage across all remaining layers, we predict a bounded \emph{lookahead window} of the next $W$ layers. This is the horizon most relevant to prefetching under limited VRAM, since experts needed much later should not compete with imminent experts for bandwidth and cache residency.
The predictor is implemented as a lightweight bottleneck MLP:
{\small
\begin{equation}
\mathbf{z}_l = \psi(\mathbf{W}_{2}\psi(\mathbf{W}_{1}\mathbf{x}_l)), \qquad
\mathbf{y}_l = \mathbf{W}_{o}\mathbf{z}_l,
\end{equation}
}
where $\psi$ denotes the BN-ReLU-Dropout block, and $\mathbf{y}_l \in \mathbb{R}^{E}$ gives an expert-priority score over the global expert pool.

To bias the predictor toward nearer layers within the lookahead window, we use a decayed supervision target. Let $\mathcal{E}_{l+d}$ denote the experts activated at layer $l+d$. For expert $e$, the target score is
{\small
\begin{equation}
g_{l,e} = \max_{1 \le d \le W} \gamma^{\,d-1}\,\mathbb{I}[e \in \mathcal{E}_{l+d}],
\end{equation}
}
where $\gamma \in (0,1]$ is a distance-decay factor. Near-future experts therefore receive higher priority than experts that appear in deeper layers of the lookahead window. At runtime, the scheduler uses the top-$B$ entries of $\mathbf{y}_l$ as an ordered prefetch candidate set: in prefill, it is a bounded priority list rather than an exact cover of future experts, whereas in decode it more directly tracks near-future hot experts.

\subsubsection{Per-layer Rolling Prediction and Runtime Integration}

The prediction module is invoked once at every MoE layer, continuously refreshing expert priorities as routing context evolves across depth. In \textbf{prefill}, rolling updates refine the ordering of a relatively broad future working set; in \textbf{decode}, they track a much sparser and more reusable set of near-future experts. At layer $l$, the module outputs an expert-priority vector $\mathbf{y}_l$, from which the runtime extracts the top-$B$ candidates for prefetch. The module requires only lightweight one-time calibration for a target VL-MoE family and is trained with soft binary cross-entropy on the distance-decayed targets $\mathbf{g}_l$. Unless otherwise stated, we use $W{=}5$, $\gamma{=}0.8$, $B{=}20$, and mean pooling for both compressed hidden-state and visual summaries. Because prediction operates on compact summary features rather than full token states, its overhead is small in practice and can be largely overlapped with ongoing execution; quantitative breakdown is reported in Section~\ref{sec:evaluation}, and implementation details are provided in ~\ref{appendix:predictor}.

\subsection{Expert Caching and Pipeline Orchestrator}
\label{subsec:optimization}

\begin{figure}[t]
\centering
    \includegraphics[width=\linewidth]{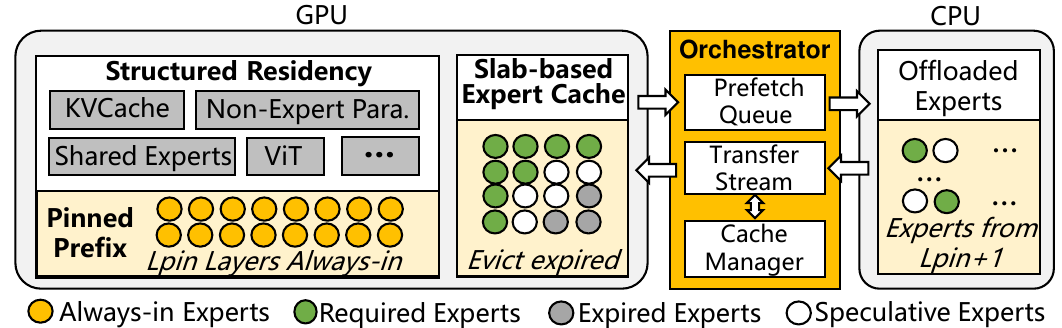}
    \caption{Data layout of \modelnamenospace.}
    \label{fig:datalayout}
\end{figure}

To support asynchronous deep prefetching under limited VRAM, we divide GPU memory into a \emph{static resident region} and a \emph{dynamic slab-based expert cache}, as shown in Figure~\ref{fig:datalayout}. The resident region keeps latency-critical parameters permanently on GPU, while the dynamic cache manages deeper experts online. Guided by the predictor, the \textbf{Expert Caching and Pipeline Orchestrator} uses near-future expert priorities to drive both prefetching and replacement.

\subsubsection{Structured Residency for Latency Masking}
\label{subsubsec:residency}

The static resident region both preserves non-swappable parameters and creates an initial overlap window in which deeper-layer transfers can proceed concurrently with computation.

\noindent\textbf{Always-resident components.}
Shared experts, which show near-100\% hit rates, remain permanently pinned. Non-expert backbone parameters (e.g., attention blocks) are likewise kept resident by the base runtime.

\noindent\textbf{Pinned prefix for routing context and startup overlap.}
We keep the \emph{entire expert pool} of the first \(L_{\text{pinned}}\) MoE layers resident on GPU. This pinned prefix serves two roles: it exposes the early routing signals required by affinity-aware compression and lookahead prediction, and it creates a startup overlap window for asynchronously prefetching deeper experts before steady-state offloading begins. Since compression takes effect only after the pinned prefix, these layers run on the uncompressed token stream. Because all experts in the prefix are resident, this stage incurs no PCIe I/O overhead, and its extra FLOPs are largely hidden by overlap with deeper-layer prefetching. Under a fixed VRAM budget, \modelname chooses a compact prefix that is sufficient for routing observability and startup overlap, leaving the remaining memory for the dynamic cache.

The prefix depth \(L_{\text{pinned}}\) is chosen at deployment time. Let \(M_{\text{avail}}\) denote the expert VRAM budget and \(S_{\text{layer}}\) the footprint of one full MoE layer. We define the valid prefix interval as
\begin{equation}
\mathcal{I}_{\text{valid}} =
\left\{
L \in \mathbb{N}
\;\middle|\;
L_{\text{semantic}} \le L \le
\left\lfloor \frac{M_{\text{avail}} - C_{\text{safe}}}{S_{\text{layer}}} \right\rfloor
\right\},
\end{equation}
where $L_{\text{semantic}}$ is the minimum depth at which routing context becomes sufficiently informative. \(C_{\text{safe}}\) is the minimum memory reserved for the dynamic cache. \ref{appendix:prefixsensitivity} provides the corresponding empirical evidence.

\subsubsection{Dynamic Slab-Based Expert Cache}

The remaining VRAM is organized as a dynamic cache for deep MoE layers (\(L > L_{\text{pinned}}\)). To avoid fragmentation and runtime allocation overhead (e.g., \texttt{cudaMalloc}), we pre-partition this space into fixed-size slabs, each storing one expert.

\noindent\textbf{Lookahead-driven insertion.}
At each MoE layer, the predictor produces a priority vector \( \mathbf{y}_l \) on GPU, which is asynchronously copied to the host. The host-side transfer stream \( \mathcal{S}_{\text{trans}} \) continuously polls this state and issues back-to-back DMA transfers for the highest-priority pending experts, avoiding synchronization stalls on the critical path.

\noindent\textbf{Safety-guarded eviction.}
When a cache slab must be reclaimed during autoregressive decoding, an expert is eligible for eviction only after its layer has completed execution for the current token. As shown in Figure~\ref{fig:datalayout}, dynamically cached experts are divided into \textbf{Required Experts}, protected within the current lookahead window; \textbf{Speculative Experts}, temporarily retained to absorb short-term prediction shifts; and \textbf{Expired Experts}, which have completed execution and are no longer protected. When space is needed, the cache manager evicts only \textbf{Expired Experts}, prioritizing those with the lowest predictor-derived priority. This policy ensures eviction safety while preserving limited prediction-guided reuse across decoding steps.

\subsubsection{I/O-Aware Pipeline Orchestration}
\label{subsubsec:io-orch}

In deep MoE offloading, PCIe transfer time typically dominates computation (\(T_{\text{trans}} \gg T_{\text{comp}}\)). Lookahead prediction therefore does not remove transfer demand, but shifts it off the critical path by turning reactive fetches into background prefetches.

We decouple execution into a transfer stream (\(\mathcal{S}_{\text{trans}}\)) and a compute stream (\(\mathcal{S}_{\text{comp}}\)). At each MoE layer, the predictor refreshes expert priorities and \(\mathcal{S}_{\text{trans}}\) fetches the highest-priority pending experts, while \(\mathcal{S}_{\text{comp}}\) executes resident layers together with the lightweight predictor. 
Affinity-aware compression reduces the effective expert working set, and the predictor concentrates bandwidth on near-future demand, allowing the transfer stream to stay ahead of computation.

If a required expert is not yet resident, \(\mathcal{S}_{\text{comp}}\) stalls only on that expert, while \(\mathcal{S}_{\text{trans}}\) continues draining the remaining queue. As a result, imperfect prediction causes localized misses rather than pipeline-wide stalls.

\section{Evaluation}
\label{sec:evaluation}

\begin{table}[t]
    \centering
    \caption{Configuration of two evaluated MMoE models.}
    \label{tab:moe_config}
    \resizebox{\linewidth}{!}{
    \begin{tabular}{lcc}
        \toprule
                                 & \textbf{Qwen3-VL-30B-A3B} & \textbf{Deepseek-VL2} \\
        \midrule
        Total Parameters         & 30B                   & 27B              \\
        Actived Parameters/Token & 3.3B                   & 4.5B             \\
        Per expert size  &  17.3 MB & 23.6 MB \\
        Total Weight Size        & 62GB                  & 55GB                          \\
        Layer Number             & 48                    & 30               \\
        Expert Number/Layer      & 128                     & 72               \\
        Activated Expert/Token              & 8                     & 6+2(shared)                \\
        \bottomrule
    \end{tabular}
    }
\end{table}

\begin{table*}[t]
\centering
\caption{Performance comparison with Vanilla (100\%) baseline and different pruning ratios (70\%, 50\%). \modelname reaches comparable accuracy with Vanilla and SOTA visual token compression methods~\cite{visionzip}.}
\label{tab:accuracy}
% 自动调整宽度以适应页面
\resizebox{\linewidth}{!}{%
\begin{tabular}{l c cc cc cc cc||}
\toprule
\multicolumn{10}{c||}{Qwen3-VL-30B-A3B} \\
\hline
\multirow{2}{*}{\textbf{Dataset}} & \textbf{Vanilla} & \multicolumn{2}{c}{\textbf{Random}} & \multicolumn{2}{c}{\textbf{Zip}} & \multicolumn{2}{c}{\textbf{Mix}} & \multicolumn{2}{c||}{\textbf{VisMMOE}} \\
% cmidrule 用于给特定列加横线，(lr) 参数让线之间有间隙，视觉分隔更明显
\cmidrule(lr){3-4} \cmidrule(lr){5-6} \cmidrule(lr){7-8} \cmidrule(lr){9-10}
 & 100\% & 70\% & 50\% & 70\% & 50\% & 70\% & 50\% & 70\% & 50\% \\
\midrule
MME      & 2498  & 2181 & 2053  & 2475 & 2309 & 2375 & 2212 & 2470 & 2299 \\
OCRBench & 829  & 674 & 554  & 819 & 741 & 793 & 685 & 801 & 705 \\
POPE     & 90.02 & 89.8 & 88.3  & 90.1 & 89.6 & 90.01& 89.6 & 90.1 & 89.6 \\
MMBench  & 86.4  & 82.6 & 80.15 & 84.8 & 83.7 & 83.3 & 82.9 & 84.6 & 83.5 \\
\bottomrule
\end{tabular}%

\begin{tabular}{l c cc cc cc cc}
\toprule
\multicolumn{10}{c}{
DeepSeek-VL2} \\
\hline
\multirow{2}{*}{\textbf{Dataset}} & \textbf{Vanilla} & \multicolumn{2}{c}{\textbf{Random}} & \multicolumn{2}{c}{\textbf{Zip}} & \multicolumn{2}{c}{\textbf{Mix}} & \multicolumn{2}{c}{\textbf{VisMMOE}} \\
% cmidrule 用于给特定列加横线，(lr) 参数让线之间有间隙，视觉分隔更明显
\cmidrule(lr){3-4} \cmidrule(lr){5-6} \cmidrule(lr){7-8} \cmidrule(lr){9-10}
 & 100\% & 70\% & 50\% & 70\% & 50\% & 70\% & 50\% & 70\% & 50\% \\
\midrule
MME      & 2243  & 2115 & 2014  & 2195 & 2142 & 2153 & 2090 & 2178 & 2103 \\
OCRBench & 803  & 631 & 591  & 780 & 685 & 761 & 646 & 771 & 673 \\
POPE     & 88.42 & 87.4 & 85.2  & 88.1 & 87.3 & 87.7& 86.9 & 88.2 & 87.3 \\
MMBench  & 77.4  & 75.1 & 72.5 & 76.6 & 75 & 76.0 & 74.1 & 76.5 & 74.7 \\
\bottomrule

\end{tabular}%
}
\end{table*}

\subsection{Experimental Setup}

\subsubsection{Hardware}
% To evaluate VisMMOE across a spectrum of memory-constrained environments, we conduct experiments on two representative platforms. First, to simulate a server-grade edge scenario where powerful compute is bottlenecked by limited VRAM, we use a workstation with a single NVIDIA A100 GPU equipped with 40GB HBM~\cite{nvidia_a100}. 
% We also experiment on a single 3090 GPU with 24GB memory. 
% These GPUs are connected to sufficient CPU memory.
% Second, for a strict embedded edge scenario, we utilize an NVIDIA Jetson AGX Orin~\cite{nvidia_jetson_orin} (32 GB shared memory). 
% For
% model weight storage, we use a Samsung 980
% PRO~\cite{samsung980pro}, which provides a theoretical read speed of 7,000
% MB/s (around 3,000 MB/s in practice). 

To evaluate VisMMOE under different memory-constrained deployment settings, we conduct experiments on three representative platforms. First, to study a server-class setting where strong compute capability is still limited by GPU memory capacity, we use a workstation with a single NVIDIA A100 GPU equipped with 40\,GB HBM~\cite{nvidia_a100}. We also evaluate on a single NVIDIA RTX 3090 GPU with 24\,GB memory, representing a more constrained commodity high-end GPU setting. In both cases, the GPUs are attached to sufficient host memory, enabling GPU/CPU offloading without host-memory bottlenecks.

Second, to explore a more constrained embedded setting, we use an NVIDIA Jetson AGX Orin~\cite{nvidia_jetson_orin} with 32\,GB shared memory. For model-weight storage, we use a Samsung 980 PRO SSD~\cite{samsung980pro}, which provides a theoretical sequential read bandwidth of 7{,}000\,MB/s and around 3{,}000\,MB/s in our practical setup. As discussed later, the Orin experiments rely on an OS-managed memory hierarchy via Linux swap, rather than a dedicated framework-level SSD offloading path or a custom direct NVMe-to-GPU paging implementation.

% By spanning from high-end workstations to power-constrained embedded devices, we demonstrate \modelnamenospace's adaptability to varying hardware limitations.

\begin{figure*}[t]
    \centering
    \subfloat[Qwen3-VL-30B-A3B on A100.]{
    \includegraphics[width=0.49\linewidth]{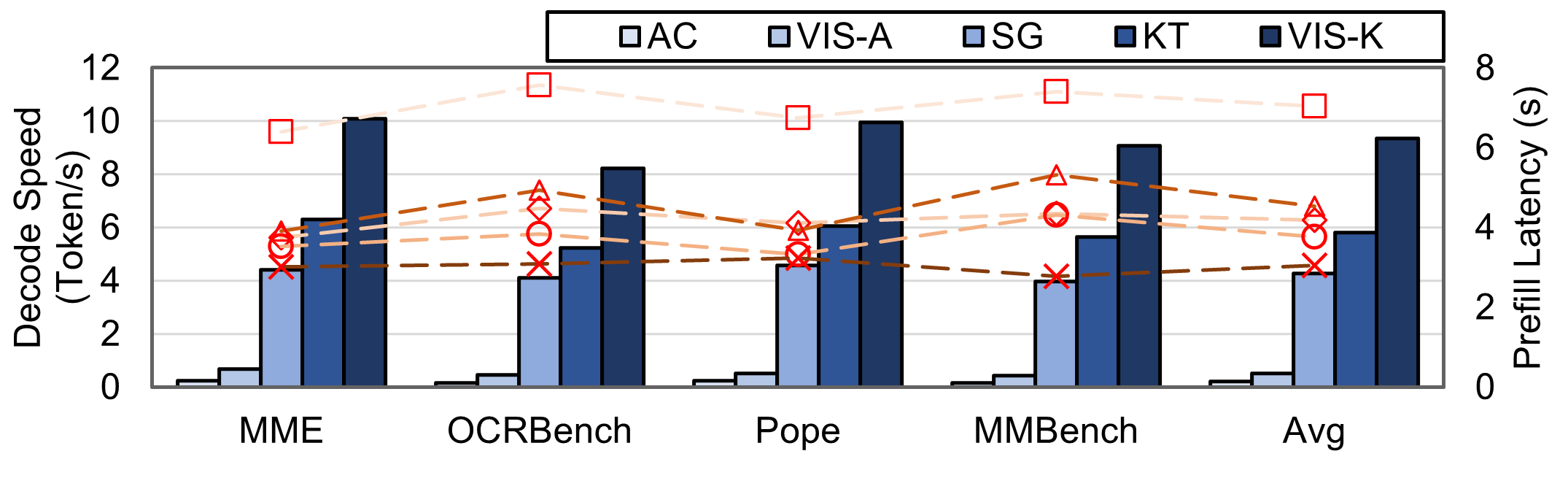}
    \label{fig:a100qwen}
    }
    \subfloat[DeepSeek-VL2 on A100.]{
    \includegraphics[width=0.49\linewidth]{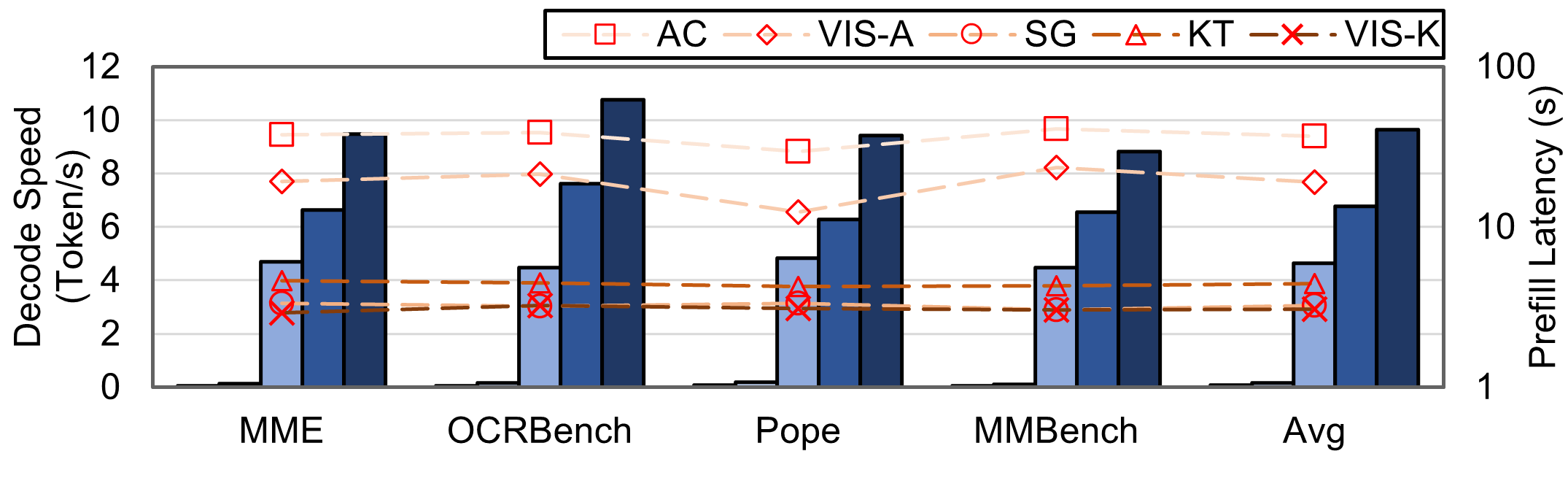}
    \label{fig:a100deepseek}
    }
    \\
    \subfloat[Qwen3-VL-30B-A3B on 3090.]{
    \includegraphics[width=0.49\linewidth]{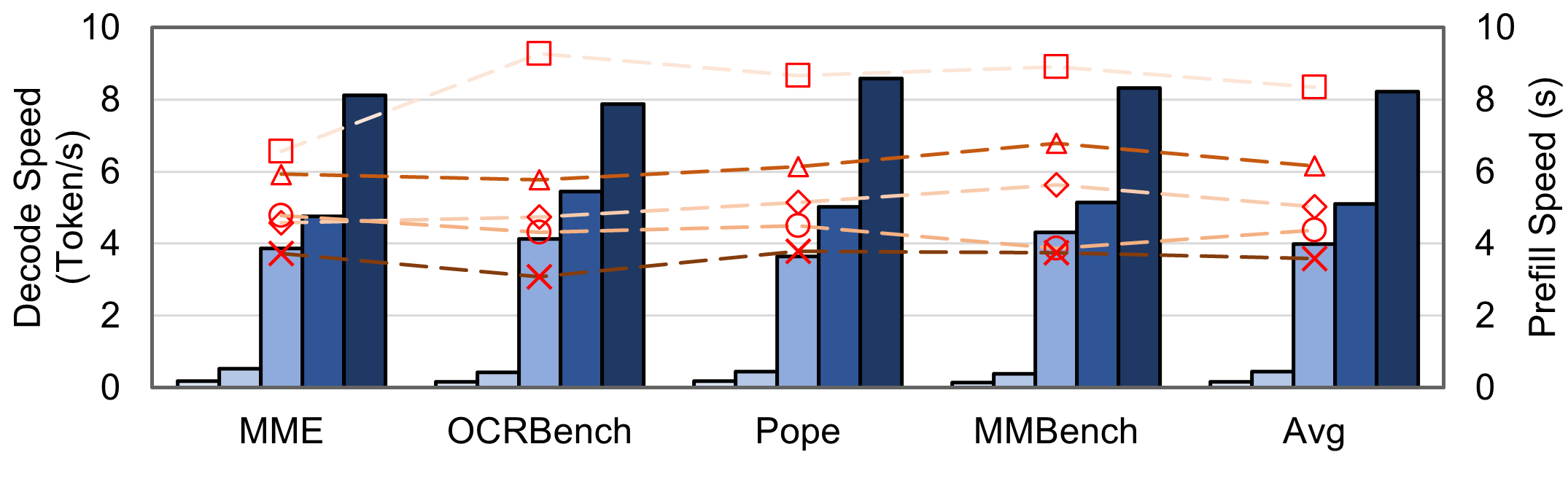}
    \label{fig:3090qwen}
    }
    \subfloat[DeepSeek-VL2 on 3090.]{
    \includegraphics[width=0.49\linewidth]{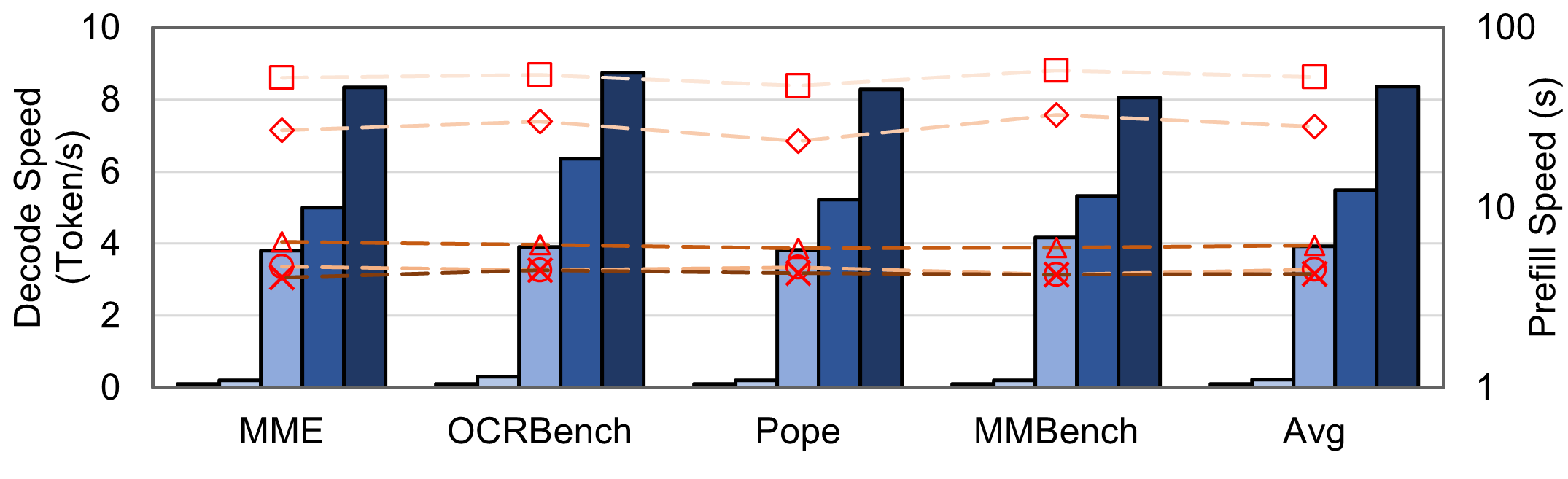}
    \label{fig:3090deepseek}
    }
    \caption{Comparison of End-to-End inference speed for \modelname and the SOTA approaches. \modelname achieves best performance on all tasks and hardware platforms in both prefill and decode stages.} 
    \label{fig:end-to-end}
\end{figure*}

\begin{figure}[t]
\centering
    \includegraphics[width=\linewidth]{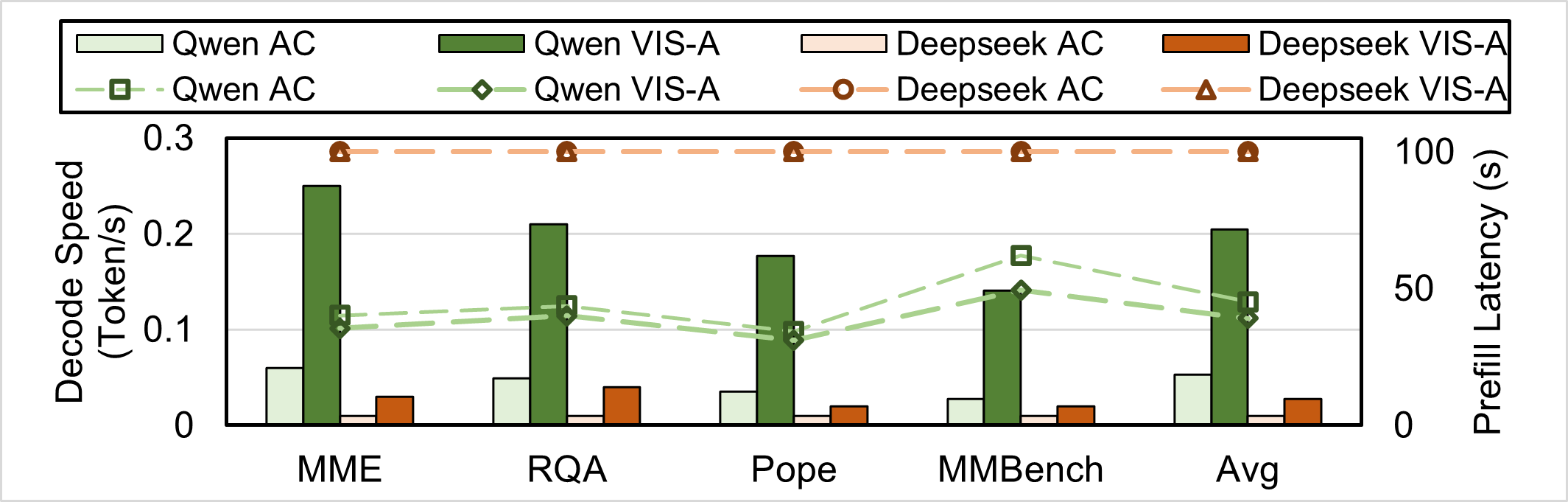}
    \caption{\modelname inference speed on Orin. Currently, sglang and ktransformers fails to support SSD swap on Orin.}
    \label{fig:orin}
\end{figure}

% \subsubsection{Implementation} 
% We implement \modelname on top of two representative frameworks: Hugging Face Accelerate~\cite{accelerate} and KTransformers~\cite{ktransformer}, modifying their memory management and computation patterns with approximately 6,000 lines of Python and C++ code. Unlike standard coarse-grained layer-wise offloading, we partition memory into a Static Resident Region and a Dynamic Expert Cache to support two distinct operating modes tailored to different hardware constraints.

% \textbf{Offload-only Mode:} The main thread queries the Deep Lookahead Predictor to asynchronously prefetch missing experts from CPU memory via the Pipeline Orchestrator. Computation on resident experts proceeds in parallel with PCIe transfers, maximizing GPU throughput by treating CPU memory strictly as a high-capacity backing store.

% \textbf{GPU-CPU Hybrid Mode:}  When the scheduler detects a cold expert that would cause a pipeline stall, it dispatches the token's intermediate activations to the CPU instead of fetching the weights. A dedicated worker thread processes these tokens using optimized AVX/AMX kernels directly in host memory, synchronizing results back to the GPU to alleviate PCIe congestion.
\subsubsection{Implementation}
We implement \modelname on top of two representative serving frameworks, Hugging Face Accelerate~\cite{accelerate} and KTransformers~\cite{ktransformer}, by modifying their memory management and execution flow with approximately 6,000 lines of Python and C++ code. Unlike conventional coarse-grained layer-wise offloading, \modelname partitions available memory into a \emph{Static Resident Region} and a \emph{Dynamic Expert Cache}, enabling different execution modes under different hardware constraints.

\textbf{Offload-only Mode.}
In this mode, offloaded experts are fetched from host memory to GPU memory on demand. The main execution thread invokes the Deep Lookahead Predictor to estimate future hot experts, while the Pipeline Orchestrator asynchronously prefetches missing experts through host-to-device transfers. Computation on already resident experts proceeds in parallel with these transfers, so host memory serves as a high-capacity backing store and the runtime focuses on maximizing compute--I/O overlap.

\textbf{GPU-CPU Hybrid Mode.} In this implementation, when the scheduler detects that a miss would stall the pipeline, it can dispatch the corresponding token activations to the CPU instead of transferring the expert weights. A worker thread executes the expert directly in host memory using optimized AVX/AMX kernels, and synchronizes the results back to the GPU. This design reduces bursty host-to-device traffic and provides a bounded fallback path under cache misses or prediction errors.

\textbf{Platform-specific deployment.}
On server-class GPUs, we use GPU/CPU offloading through the host-memory-based execution path supported by the underlying frameworks. On Jetson Orin, because no mature framework-level SSD offloading path is available for the evaluated VL-MoE models, we rely on Linux swap to extend effective host memory capacity onto SSD. Accordingly, the Orin results reflect an OS-managed memory hierarchy rather than a custom direct NVMe-to-GPU paging design.

\subsubsection{Models}
We select two representative mainstream MoE models in Table~\ref{tab:moe_config} that exhibit substantial differences in overall scale, architectural design, and MoE
layer implementation. This diverse selection enables a more
comprehensive and rigorous evaluation of our system’s scalability and generalization capability across heterogeneous configurations.

\subsubsection{Datasets}
To rigorously evaluate \modelname across diverse modalities and difficulty levels, we select four representative benchmarks: MME~\cite{mme} and MMBench-EN~\cite{mmbench} for comprehensive multi-modal perception and reasoning; POPE~\cite{pope} for evaluating object hallucination and fine-grained visual consistency; OCRBench~\cite{ocrbench} for OCR capabilities. 
% These benchmarks collectively demonstrate that
% pruned inferences maintain remarkably high accuracy across a wide spectrum of scenarios.

\subsubsection{Baselines}
We evaluate our system against three representative baselines. The first is the hugging face accelerate~\cite{accelerate} (AC), a general LLM library providing basic support for transformers. The second is the sglang~\cite{sglang} (SG), a high-performance serving framework for large language models. The third is the ktransformer~\cite{ktransformer} (KT), a CPU-optimized framework for heterogeneous LLM inference which specifically optimizes MOE offloading. We built \modelname both on accelerate and ktransformer, creating two versions of \modelname (VIS-A and VIS-K).  
All experiments are conducted under 50\% pruning ratio.

Currently, prior academic MoE offloading systems such as MoE-Infinity~\cite{moeinfinity}, MoE-Offloading~\cite{moe-offloading}, and HOBBIT~\cite{hobbit} are primarily designed for text-centric MoE workloads and do not provide support for the VL-MoE models evaluated in this paper. As a result, we do not include them in our experimental comparison. We therefore limit our empirical claims to the baselines that can be run reliably on the target VL-MoE models and hardware platforms. 
% A direct comparison with these MoE-specific offloading systems under a unified VL-MoE implementation would be valuable future work.

% We evaluate our system against both practical baselines and an approximation of prior academic MoE offloading designs. The practical baselines are Hugging Face Accelerate~\cite{accelerate} (AC), SGLang~\cite{sglang} (SG), and KTransformers~\cite{ktransformer} (KT). We build \modelname on top of both Accelerate and KTransformers, yielding VIS-A and VIS-K. All experiments are conducted under a 50\% pruning ratio.

% To additionally compare against prior academic MoE offloading ideas, we implement a MoE-Infinity-style~\cite{moeinfinity} approximation within the same VL-MoE runtime. This approximation preserves the key ideas of request-level activation tracing, activation-aware cache management, and trace-guided prefetching, while using the same model, backend, and hardware stack as \modelname. We take this approach because prior systems such as MoE-Infinity~\cite{moeinfinity}, MoE-Offloading~\cite{moe-offloading}, and HOBBIT~\cite{hobbit} target text-centric MoE workloads and do not provide runnable support for the VL-MoE models studied here. Implementing their core ideas within the same runtime enables a fairer comparison by isolating algorithmic differences from framework porting effects.

\subsection{Comparison with SOTA}

\subsubsection{Model Accuracy}
Since we prune less important tokens, it is crucial to verify that \modelname does not sacrifice performance. 
Table~\ref{tab:accuracy} reports the performance comparison between VisMMOE and various baselines, including full-loading and other pruning strategies, across different compression ratios. VisMMOE achieves accuracy comparable to state-of-the-art token compression methods (e.g., VisionZip) while significantly reducing the token count. Compared with the Vanilla (100\%) accuracy baseline, VisMMOE (70\%) only shows 1.12\%, 3.38\%, -0.09\% (gain) and 2.08\% of accuracy downgrade respectively for \textit{MME}, \textit{OCRBench}, \textit{POPE} and \textit{MMBench} on Qwen3-VL, suggesting that our affinity-aware pruning may effectively filter out visual routing noise. 
The only noticeable performance gap occurs in OCRBench when we prune 50\% of the visual tokens, which is highly sensitive to high-resolution visual details. While a slight drop is observed, VisMMOE still significantly outperforms random pruning and remains competitive with SOTA token compression methods. 
This confirms that VisMMOE effectively balances system efficiency with reasoning fidelity.

\subsubsection{End-to-end Performance}
From  Figure~\ref{fig:a100qwen} and Figure~\ref{fig:a100deepseek}, we can observe that \modelname delivers the best performance in terms of both decoding speed and prefill latency for the evaluated models. The \modelname implementation based on Accelerate~\cite{accelerate} performs 2.68x and 2.29x better in decode speed, and possess 52\% and 58.7\% of prefill latency respectively for Qwen and Deepseek-VL2. Note that the huggingface implementation of Deepseek-VL2 offload is extremely inefficient as it is based on old frameworks (torch 2.0.1)  with customized implementation. For ktransformer-based~\cite{ktransformer} \modelname implementation, Vis-K also performs 1.61x and 1.42x better in decode speed. VIS-K achieves 34\% and 31\% of prefill latency reduction. Ktransformer performs poor in prefill stage since it has to compute large proportion of experts on CPU (all experts are involved in prefilling while only 6-8 in decode). In contrast, \modelname tries to compute hot experts on GPU. Vis-K also significantly outperforms sglang~\cite{sglang}, which is a widely used serving framework.

Similar conclusions can be drawn on RTX 3090 as shown in Figure~\ref{fig:3090qwen} and~\ref{fig:3090deepseek}. The \modelname implementation on accelerate achieves up to  2.53x decode speedup and 47\% of prefill reduction compared with accelerate, while the \modelname implementation on ktransformer achieves up to 1.54x decode speedup and 37\% of prefill reduction. 

Additionally, we experiment on Jetson Orin as shown in Figure~\ref{fig:orin}. Unfortunately, sglang~\cite{sglang} and ktransformers~\cite{ktransformer} fail to support SSD swap on Orin with 32GB unified memory. They suffer severe \textit{NvMapMemAllocInternalTagged} error in load weight stage.  While VIS-A actually outperforms accelerate~\cite{accelerate} more on Jetson due to the low SSD bandwidth in swapping, the extremely low decode speed and high prefill latency prevents it from practical usage. 

\subsection{Ablation Study}

% \begin{table}[t]
% \caption{Ablation Study Results.}
% \label{tab:ablation}
%     \centering
%     \resizebox{\linewidth}{!}{
%     \begin{tabular}{clcc} % c=居中, l=左对齐
%         \toprule
%         \multicolumn{4}{c}{ Qwen3-VL-30B-A3B}\\
%         \midrule
%         \midrule
%         Implementation & Method & Speed (s) & Norm. \\
%         \midrule
%         \multirow{4}{*}{\shortstack{Huggingface \\Transformer}} 
%               & Base (Pure transformer) & 13.17 & 100\% \\
%               & + compression       & 11.55($-$1.62) & 87.7\% \\
%               & + prediction \& prefetching         & 7.72 ($-$5.45) & 58.6\% \\
%               & + optimization (VisMMOE)             & 4.91 ($-$8.26) & 37.2\% \\
%         \bottomrule
%         \midrule
%         Implementation & Method & Speed (s) & Norm. \\
%         \midrule
%         \multirow{4}{*}{\shortstack{Ktransformer}} 
%               & Base (Pure Ktransformer) & 3.89 & 100\% \\
%               & + compression       & 3.71($-$0.18) & 95.3\% \\
%               & + prediction \& prefetching         & 2.34($-$1.55) & 60.1\% \\
%               & + optimization (VisMMOE)             & 2.21 ($-$1.68) & 56.8\% \\
%         \bottomrule
%         \midrule
%         Memory Breakdown & Anchor size & Static size & Cache size \\
%         \midrule
%         Total:40GB & 21.6GB & 4.1GB & 14.3GB  \\ 
%         \bottomrule
%     \end{tabular}
%     }
    
% \end{table}

\begin{table}[t]
\caption{Ablation: speed and memory budget breakdown.}
\label{tab:ablation}
\centering
\resizebox{\linewidth}{!}{
\begin{tabular}{clcc}
\toprule
\multicolumn{4}{c}{Qwen3-VL-30B-A3B} \\
\midrule
Implementation & Method & Speed (s) & Norm. \\
\midrule
\multirow{4}{*}{\shortstack{Hugging Face\\Transformers}}
  & Base (pure transformer)            & 13.17         & 100\% \\
  & + compression                      & 11.55 (-1.62) & 87.7\% \\
  & + prediction \& prefetching        & 7.72 (-5.45)  & 58.6\% \\
  & + optimization (VisMMOE)           & 4.91 (-8.26)  & 37.2\% \\
\midrule
\multirow{4}{*}{KTransformers}
  & Base (pure KTransformers)          & 3.89          & 100\% \\
  & + compression                      & 3.71 (-0.18)  & 95.3\% \\
  & + prediction \& prefetching        & 2.34 (-1.55)  & 60.1\% \\
  & + optimization (VisMMOE)           & 2.21 (-1.68)  & 56.8\% \\
\midrule
\midrule
A100-40GB memory & anchor: 21.6\,GB & static: 4.1\,GB & cache: 14.3\,GB \\
\bottomrule
\end{tabular}
}
\end{table}

\subsubsection{Speedup Breakdown}
The ablation study in Table~\ref{tab:ablation} demonstrates the individual and combined impact of three key optimizations in VisMMOE: affinity-aware compression, prediction-guided prefetching, and pipeline optimization. When all optimizations are enabled, the system achieves a latency of 4.91s (37.2\% normalized) on the Huggingface backend. Disabling the deep prediction and pipeline optimization (i.e., using only compression) increases latency to 11.55s (87.7\%), indicating a modest but measurable benefit derived solely from I/O volume reduction. In contrast, enabling prediction and prefetching leads to a more significant drop to 7.72s (58.6\%), highlighting its substantial contribution to performance by effectively masking the massive expert transfer overhead. When all optimizations are removed, performance further degrades to 13.17s. Notably, the performance gains from stacking these components are cumulative, suggesting that compression acts as an enabler for the predictor, while the pipeline orchestrator ensures the theoretical bandwidth gains are realized in practice.

\modelname also achieves 1.76x speedup on ktransformers. Compression  show limited effects as ktransformer is already well-optimized and I/O far overweigh computation. Most of the benefits comes from predicting and placing hot experts on GPU both in prefill and decode stages.  

For Qwen3-VL-30B-A3B on A100-40GB, VRAM is divided among the backbone, pinned anchor layers, and the dynamic expert cache. The backbone occupies 4.1\,GB, anchor layers take 21.6\,GB, and the remaining 14.3\,GB is used for the cache. This illustrates the tradeoff in VisMMOE: deeper anchoring improves early routing visibility for prediction, but reduces cache capacity for later-layer experts. On RTX 3090, the tighter VRAM budget allows us to pin only the first four layers, which weakens prediction and shortens the effective prefetch window. As reflected in our main results, this leads to smaller end-to-end gains on the 3090 than on the A100.

% Furthermore, our system achieves a significantly higher speedup ratio on the unoptimized Huggingface backend (2.68$\times$) compared to the highly optimized Ktransformer backend (1.85$\times$). This consistent gain across disparately optimized runtimes further amplifies the performance benefit of our I/O-centric design. 

\begin{figure}[t]
\centering
    \includegraphics[width=\linewidth]{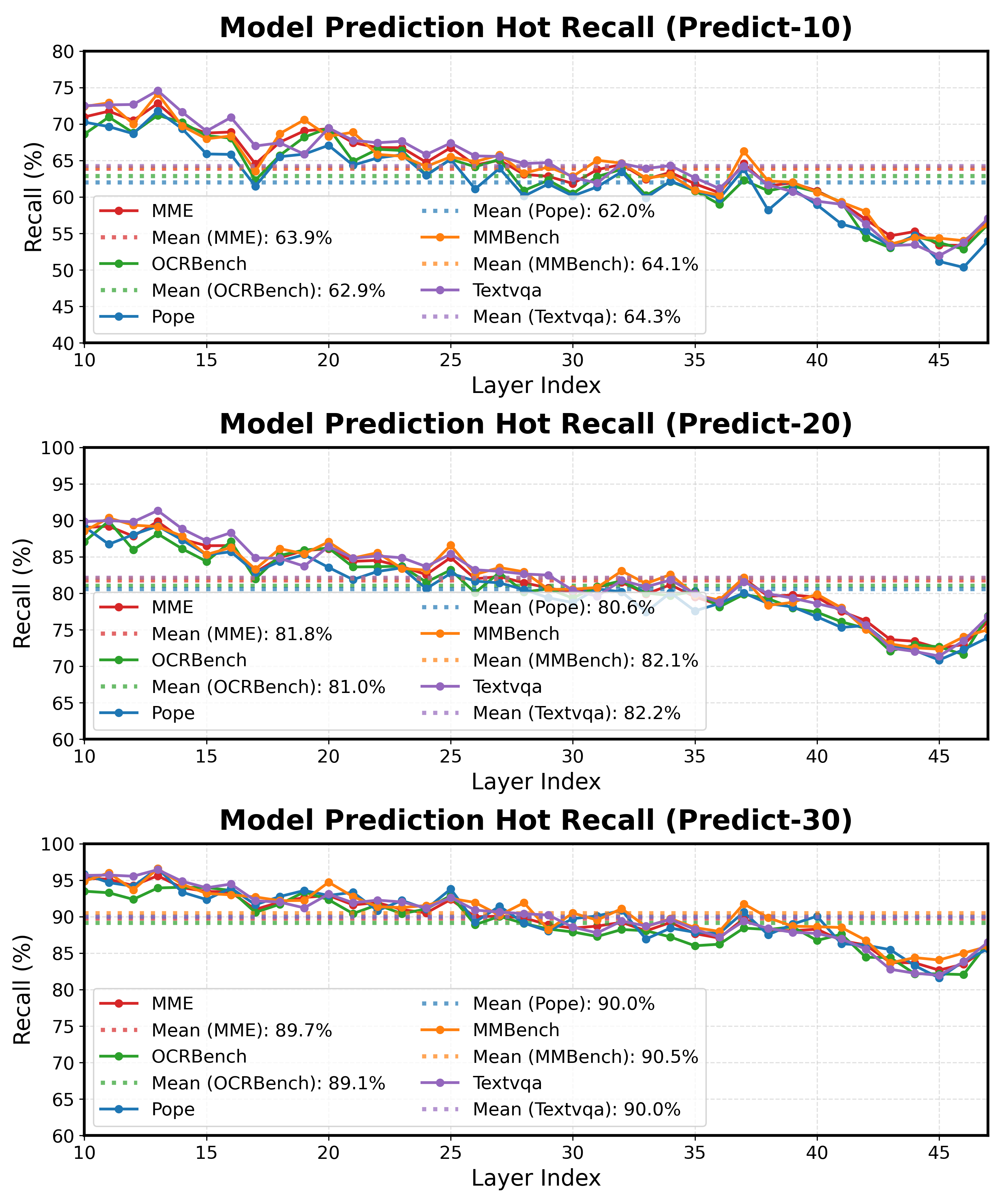}
    \caption{Layer-wise \textit{Hot Recall} performance of the expert predictor on different datasets. The proposed predictor significantly outperforms the random baselines by 3.82x (predict-30) to 7.9x (predict-10). Note that the \modelname predictor works well even for \textit{textvqa}, which does not participate in training at all. Layers 0-9 are pinned on the GPU memory and do not require prediction.}
    \label{fig:prediction_recall}
\end{figure}

% \begin{figure}[t]
%     \centering
%     % Subfigure A
%     \subfloat[Model recall prediction 10 experts.]{
%         \includegraphics[width=\linewidth]{figures/plot_recall_5_datasets_k10.png}
%         \label{fig:recall10}
%     } \\ % <-- Added newline to stack them vertically
    
%     % Subfigure B
%     \subfloat[Model recall prediction 20 experts.]{
%         \includegraphics[width=\linewidth]{figures/plot_recall_5_datasets_k20.png}
%         \label{fig:recall20}
%     } \\ % <-- Added newline to stack them vertically
    
%     % Subfigure C
%     \subfloat[Model recall prediction 30 experts.]{
%         \includegraphics[width=\linewidth]{figures/plot_recall_5_datasets_k30.png}
%         \label{fig:recall30}
%     }
    
%     % Updated caption to match the actual recall graphs
%     \caption{Comparison of model prediction hot recall for \modelname across different datasets. The subfigures demonstrate the recall performance across evaluated layers for top-10, top-20, and top-30 expert predictions.} 
%     \label{fig:recall}
% \end{figure}

\begin{figure}[t]
    \centering
    % 左边的独立图
    \begin{minipage}{0.48\columnwidth}
        \centering
        \includegraphics[width=\linewidth]{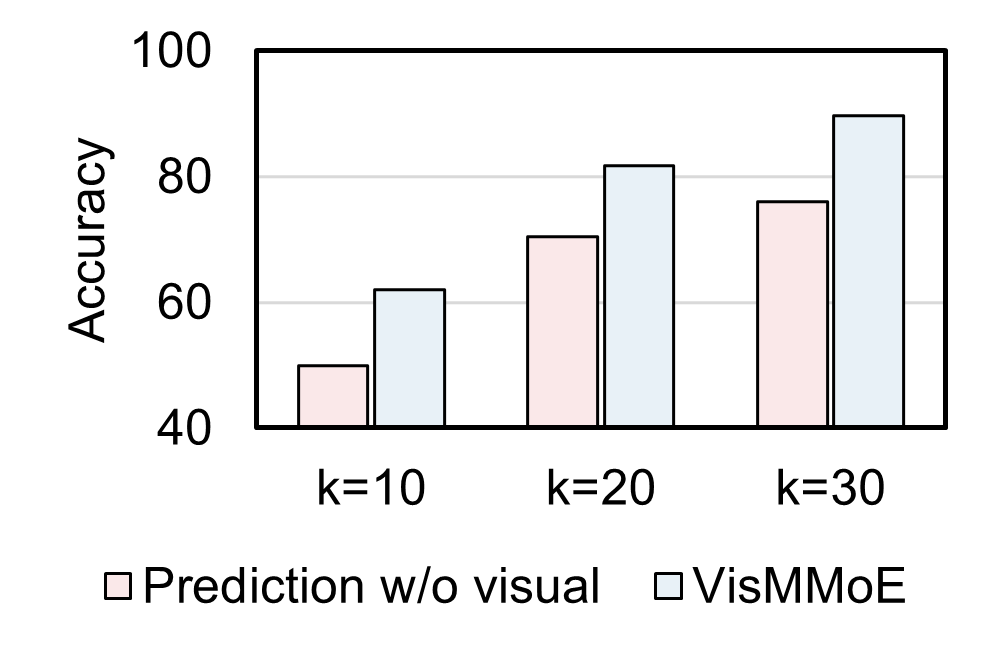}
        \caption{The effect of visual info in prediction.}
        \label{fig:predictionablation}
    \end{minipage}\hfill
    % 右边的独立图
    \begin{minipage}{0.48\columnwidth}
        \centering
        \includegraphics[width=\linewidth]{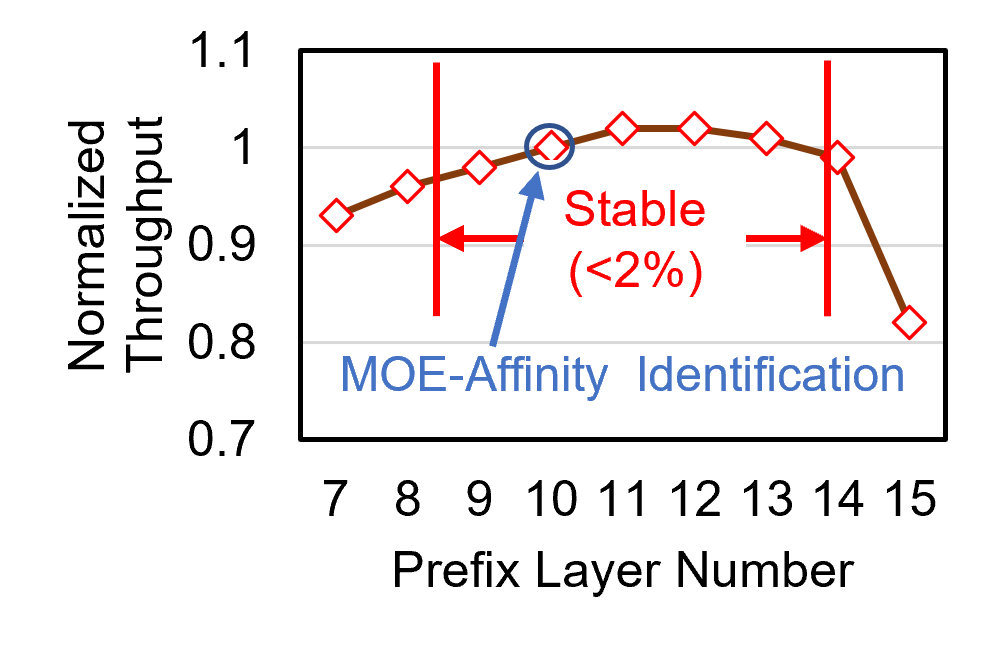}
        \caption{Anchor layer number choice and effect.}
        \label{fig:anchorablation}
    \end{minipage}
\end{figure}

\begin{figure}[t]
\centering
    \includegraphics[width=\linewidth]{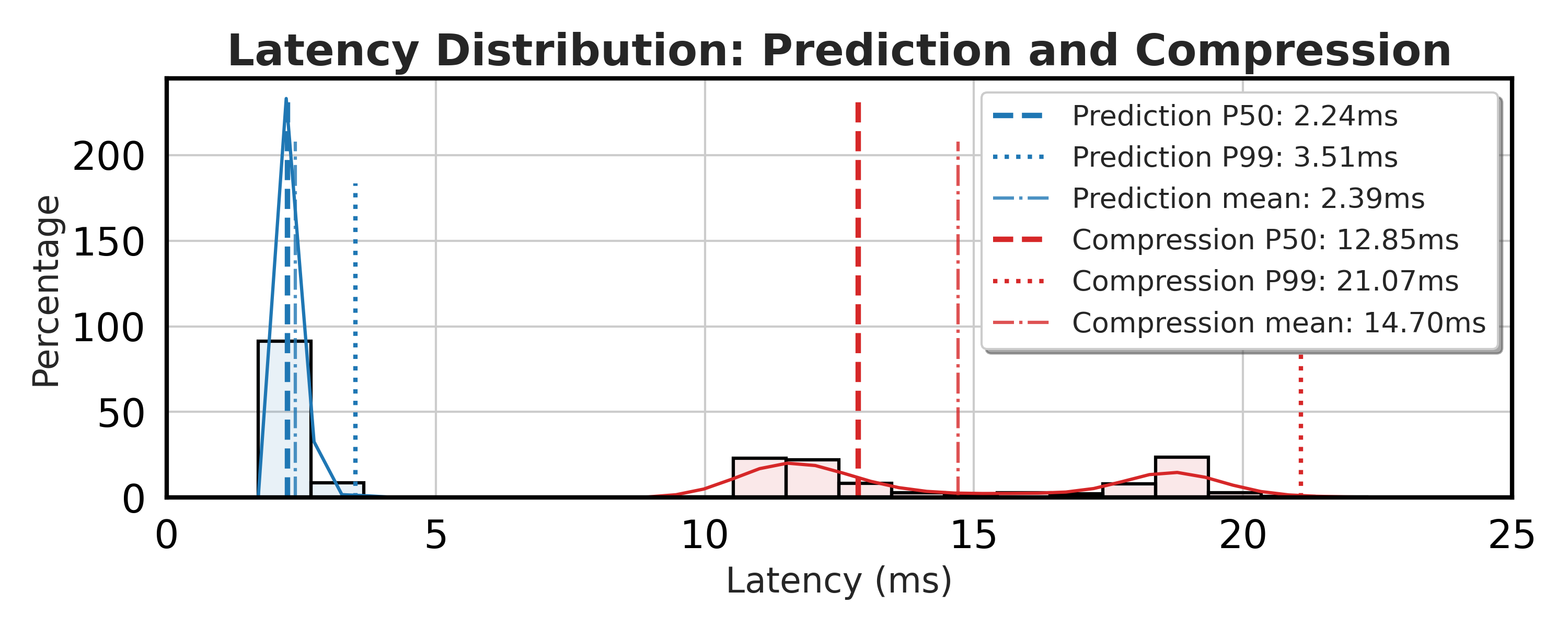}
    \caption{\modelname compression and predictor costs. Presented in latency distribution.}
    \label{fig:predictorlatency}
\end{figure}

\subsubsection{Prediction Accuracy}
We evaluate the effectiveness of our Predictor using Hot Recall, the ratio of activated experts successfully prefetched into the cache. As illustrated in Figure~\ref{fig:prediction_recall}, the predictor significantly outperforms the random baseline, which yields a meager recall of 7.8\% ($k=10$) to 23.4\% ($k=30$). In stark contrast, VisMMOE achieves 64.1\%, 81.9\%, and 89.8\% for buffer sizes of $k=10, 20, 30$ respectively on five different datasets. This represents a massive 3.8$\times$ to 8.2$\times$ improvement, confirming that our affinity-aware visual compression successfully exposes predictable routing patterns. While the predictor is trained on partial data of \textit{MME, OCRBench, MMBench and Pope}, it also shows similar effect on \textit{textvqa} which does not participate in training at all. It proves the versatility‌ of our predictor.

% Furthermore, the predictor demonstrates exceptional robustness against error accumulation in deep layers. Thanks to our Rolling Horizon Update mechanism which continuously refines forecasts with fresh semantic states, recall remains consistently high ($>$85\% at $k=30$) even at Layer 30+. This high accuracy is system-critical: the vast majority of expert requests are served directly from the high-speed GPU cache, allowing the pipeline orchestrator to effectively hide the PCIe latency of the few remaining misses.

\subsubsection{Effect of visual information on expert prediction.}
Figure~\ref{fig:predictionablation} highlights the importance of visual information in expert prediction. We compare the full predictor with an already competitive ablated variant that relies only on hidden states and historical routing traces, without the ViT visual feature ($h_l^{(v)}$). This ablated predictor is substantially stronger than a random or history-free baseline, since it still captures evolving textual states and prior expert activation patterns. Nevertheless, its Hot Recall drops markedly across all prediction horizons once visual features are removed. 
% This shows that our predictor's advantage is not merely due to better exploitation of textual dynamics, but also critically depends on compressed visual representations as a stable semantic anchor for deep lookahead.

\subsubsection{Anchor layer number choice and effect.}
Figure~\ref{fig:anchorablation} plots the end-to-end execution time across different $L_{\text{pinned}}$ values. The results reveal a clear U-shaped performance curve with a distinct, stable optimal region (e.g., $L_{\text{pinned}} \in [8, 12]$ for Qwen3-VL). \modelnamenospace's offline profiling successfully identifies and selects an anchor size within this region. This empirical analysis confirms that \modelnamenospace's offline selection effectively balances prediction accuracy, initial latency masking, and steady-state cache stability.

\subsubsection{Prediction/Compression Cost}
To assess the runtime cost of our introduced modules, we profile the latency distribution of the Affinity-Aware Token Compression and the Compression-Guided Predictor, as illustrated in Figure~\ref{fig:predictorlatency}.
The predictor proves to be extremely lightweight, introducing a negligible overhead with a mean latency of 2.72 ms (P50: 2.78 ms, P99: 3.15 ms). This sub-3ms latency is orders of magnitude smaller than the expert transfer time (tens of ms), ensuring that the lookahead mechanism essentially runs for free alongside the main computation. The compression module incurs a moderate one-time cost, averaging 14.70 ms (P50: 12.85 ms, P99: 21.07 ms). While higher than prediction, this overhead is strategically masked. As detailed in Section 3.4, compression is performed only once during the execution of the resident  layers. 
% The 14.7 ms latency is fully absorbed by the initial prefetching window for deep layers, preventing it from exposing  stalls on the end-to-end critical path.

% \begin{figure}[t]
%   \setlength{\belowcaptionskip}{-10pt}
%     \subfloat[Random prefetching.]{
%     \includegraphics[width=\linewidth]{figures/plot_random_baselines_zoomed.png}
%     \label{fig:recallrandom}
%     }
%     \\
%      \subfloat[Prediction-based prefetching.]{
%     \includegraphics[width=\linewidth]{figures/plot_model_recall_means.png}
%     \label{fig:recallprediction}
%     }
%     % \subfloat[Cost breakdown of the in-sensor computation.]{
%     % \includegraphics[width=0.35\linewidth]{figures/isca-bg3.pdf}
%     % \label{fig:bg2} }
%     \caption{Layer-wise \textit{Hot Recall} performance of the expert predictor. The proposed predictor significantly outperforms the random baseline. Layers 0-9 are pinned on the GPU memory and do not require prediction.} 
%     \label{fig:prediction}
% \end{figure}

\section{Related Works}
\label{sec:relatedworks}

% \textbf{Visual Token Compression:} 
% While Multi-Modal Large Language Models demonstrate exceptional capabilities in processing diverse data modalities, they impose substantial computational burdens. Beyond standard techniques like quantization and pruning, token compression has emerged as a critical optimization strategy, particularly for handling high-resolution visual inputs~\cite{survey}. Existing methodologies, exemplified by VisionZip~\cite{visionzip}, predominantly focus on identifying \textit{intra-image redundancy}. These approaches~\cite{divprune,fastvid,hiprune} typically rely on intrinsic model signals to identify and retain semantically salient patches. However, these saliency-driven methods~\cite{LLaVA-Scissor,llavamini, PACT, videochat,similarity,visionselector} operate primarily at the algorithmic level, often neglecting the implications for system execution. \textit{In contrast, \modelname systematically investigates the interplay between token compression and network execution to maximize actual performance savings.}

\textbf{Visual Token Compression:}
Visual token compression has become an important optimization  under high-resolution images and long visual sequences~\cite{survey}. Existing methods~\cite{PACT,videochat,similarity,visionselector}, such as VisionZip~\cite{visionzip}, mainly remove \textit{intra-image redundancy} by selecting salient or non-redundant patches using saliency signals, similarity metrics, or learned policies~\cite{visionzip,divprune,fastvid,hiprune,LLaVA-Scissor,llavamini}. FastMMOE~\cite{xia2025fastmmoeacceleratingmultimodallarge} first take routing into consideration in token compression. However, these methods mainly optimize algorithmic efficiency, such as reducing FLOPs or KV-cache growth, without explicitly considering downstream expert offloading behavior. \textit{In contrast, \modelname uses visual token compression as a systems lever to compact the expert working set and improve expert predictability during offloaded VL-MoE inference.}

\textbf{General LLM Inference Systems:}
A large body of work improves LLM serving efficiency through better memory management, batching, and operator optimization. Systems such as vLLM~\cite{vllm} and DeepSpeed~\cite{deepspeed} improve throughput and latency through optimized runtime support, while Accelerate~\cite{accelerate} and llama.cpp~\cite{llama} enable deployment on memory-constrained devices via CPU or NVMe offloading. Other systems, including Deja Vu~\cite{DejaVu}, PowerInfer~\cite{powerinfer}, and CATS~\cite{cats}, exploit activation sparsity in dense models to skip redundant computation. However, these systems mainly target dense models or general-purpose serving, rather than the dynamic expert movement bottleneck in sparse MoE inference. \textit{In contrast, \modelname targets offloaded VL-MoE inference and optimizes expert locality, prefetchability, and cache efficiency under limited GPU memory.}

\textbf{Expert Offloading for MoE-based Models:}
Prior work reduces the memory footprint of MoE models by streaming experts on demand. Systems such as MoE-Offloading~\cite{moe-offloading}, MoE-Infinity~\cite{moeinfinity}, and SwapMoE~\cite{swapmoe} hide I/O latency through expert prefetching and cache management, while KTransformers~\cite{ktransformer} and Fiddler~\cite{fiddler} further leverage heterogeneous CPU-GPU execution. However, these systems mainly optimize expert movement under a given routing pattern and are largely evaluated on text-centric workloads, without explicitly exploiting the routing characteristics introduced by visual tokens. \textit{In contrast, \modelname is complementary to this line of work: it improves the effectiveness of expert caching and prefetching in visual-heavy VL-MoE settings by making expert demand itself more compact and predictable.}

\section{Conclusion}
\label{sec:conclusion}
% In this work, we introduce \modelnamenospace, a holistic and efficient inference system for deploying large-scale Multi-modal MoE  models on memory-constrained edge devices. By addressing the high latency and bandwidth bottlenecks associated with random expert loading, \modelname exploits Visual-Expert Affinity to regularize access patterns, enabling significant speedups in inference performance. Additionally, we have implemented \modelname atop both the Hugging Face Accelerate and KTransformer frameworks. 
% % This dual-implementation strategy ensures flexible scalability and compatibility across diverse software stacks. 
% Overall, \modelname represents a significant step forward in democratizing massive multi-modal models.

In this work, we present \modelnamenospace, a system for efficient inference of large-scale vision-language MoE models on memory-constrained single-GPU platforms. Our central insight is that visual token compression can be used not only to reduce visual processing cost, but also to compact the expert working set and improve expert predictability during offloaded inference. Guided by this observation, \modelname reduces the latency and bandwidth overhead of dynamic expert loading through the co-design of compression, prediction, and runtime scheduling. We implement and validate \modelname on top of multiple software backends. More broadly, our results suggest that co-optimizing model inputs and system execution is a promising direction for practical VL-MoE deployment under tight memory budgets.

\bibliographystyle{ACM-Reference-Format}
\bibliography{main}

\end{document}